%% file: eacl2021.tex
\documentclass[11pt,a4paper]{article}
\usepackage[hyperref]{eacl2021}
\usepackage{times}
\usepackage{latexsym}

\usepackage{microtype}

\usepackage{amsmath}
\usepackage{appendix}
\usepackage{booktabs}
\usepackage{caption}
\usepackage{color}
\usepackage{dblfloatfix}
\usepackage{diagbox}
\usepackage{graphbox}
\usepackage{graphicx}
\usepackage{makecell}
\usepackage{multirow}
\usepackage{url}
\usepackage{subfig}
\usepackage{subfiles}
\usepackage{multirow}
\usepackage[normalem]{ulem} 
\usepackage{xspace}

\usepackage{natbib}
\usepackage{enumitem}
\usepackage{textcomp}

\aclfinalcopy 
\def\aclpaperid{1094} 

\newcommand{\DEit}{DE\textsubscript{I2T}\xspace}
\newcommand{\DEtt}{DE\textsubscript{T2T}\xspace}

\newcommand{\DEttit}{DE\textsubscript{T2T+I2T}\xspace}

\title{Crisscrossed Captions: Extended Intramodal and Intermodal Semantic Similarity Judgments for MS-COCO}
\author {
    Zarana Parekh, Jason Baldridge, Daniel Cer, Austin Waters, and Yinfei Yang \\
    Google Research \\
    \texttt{\{zarana, jasonbaldridge, cer, austinwaters, yinfeiy\}@google.com}
}

\date{}

\begin{document}

\maketitle

\begin{abstract}
    By supporting multi-modal retrieval training and evaluation, image captioning datasets have spurred remarkable progress on representation learning. Unfortunately, datasets have limited cross-modal associations: images are not paired with other images, captions are only paired with other captions of the same image, there are no negative associations and there are missing positive cross-modal associations. This undermines research into how inter-modality learning impacts intra-modality tasks. We address this gap with \textit{Crisscrossed Captions} (CxC), an extension of the MS-COCO dataset with human semantic similarity judgments for \textbf{267,095} intra- and inter-modality pairs. We report baseline results on CxC for strong existing unimodal and multimodal models. We also evaluate a multitask dual encoder trained on both image-caption and caption-caption pairs that crucially demonstrates CxC's value for measuring the influence of intra- and inter-modality learning.
\end{abstract}

\section{Introduction}
\begin{figure}[t]
\begin{center}
\includegraphics[width=1\linewidth]{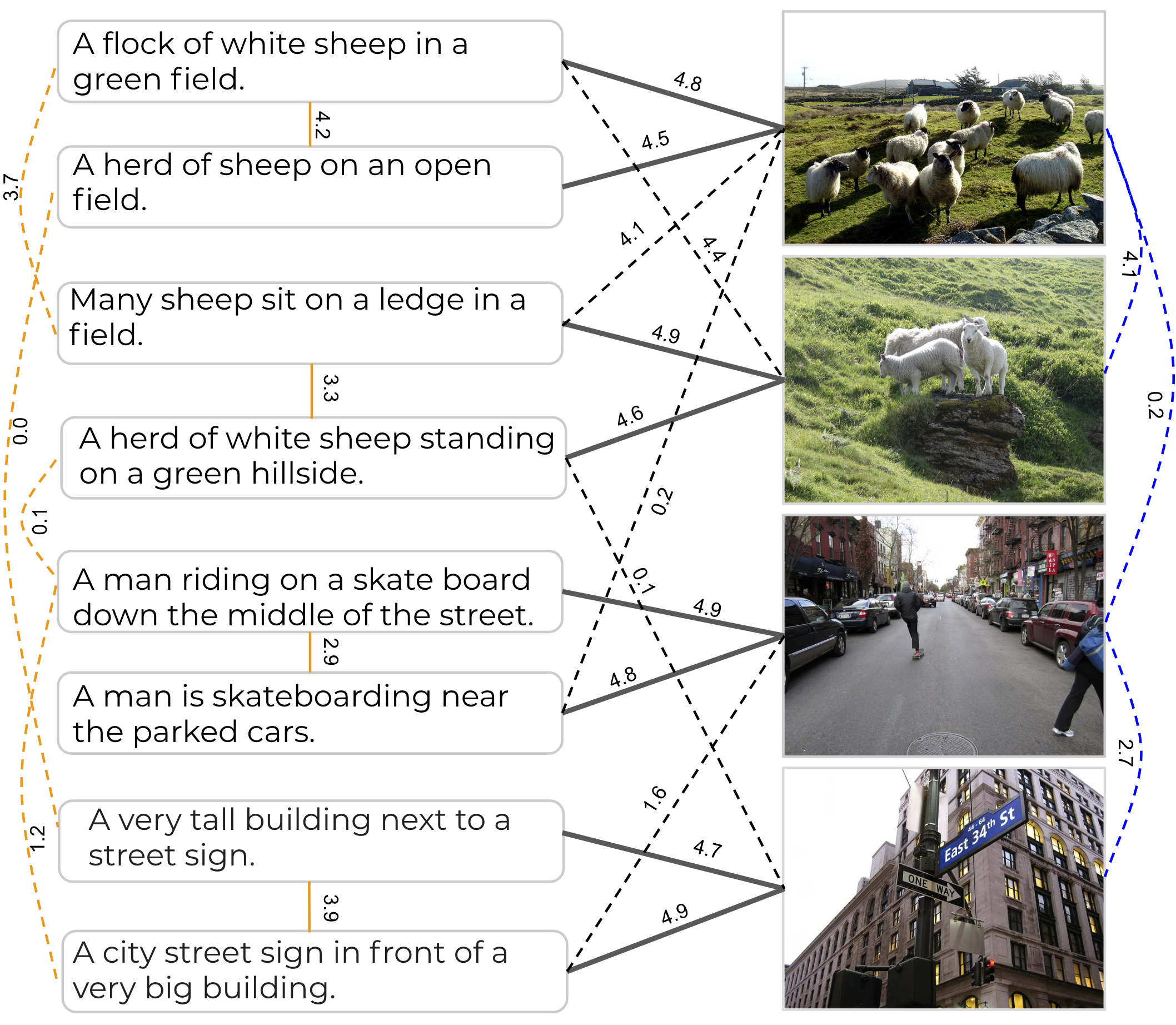}
\end{center}
\caption{Crisscrossed Captions extends the MS-COCO evaluation sets by adding semantic similarity ratings for existing image-caption pairs and co-captions (solid lines), and it increases annotation density by adding further ratings for new image-caption, caption-caption and image-image pairs (dashed lines).}
\label{fig:overview}
\end{figure} 


Phrases such as \textit{blue}, \textit{chair}, and \textit{garden path} have strong visual components, yet computational word representations are usually created with text-only corpora. Encouragingly, some recent work that derives representations using visual contexts shows improvements for both word similarity ranking and image-text retrieval \cite{kiros2018illustrative}, and query-based training of image models demonstrates language's power to improve image representations \cite{graph-rise}. Learning representations for both vision and language jointly should be even more effective---indeed, much progress has been made on such cross-modal learning using image captioning data \cite{karpathy2015deep,harwath-glass-2017-learning,DBLP:conf/bmvc/FaghriFKF18,Li_2019_ICCV}. However, it is not yet clear whether learning representations in multimodal contexts improves performance \textit{within} as well as across modalities as there are no datasets ideally suited for this at present.

Image captioning datasets such as Flickr8k \cite{rashtchian2010collecting}, Flickr30k \cite{young2014image}, Multi30k \cite{elliott-etal-2016-multi30k}, Microsoft Common Objects in COntext (MS-COCO) \cite{lin2014microsoft}, and Conceptual Captions \cite{sharma2018conceptual} only capture relationships between images and textual captions created for them. They miss many valid relationships between unassociated images and captions, from captions to other captions, and from images to other images. We address this gap with Crisscrossed Captions (CxC, exemplified in Figure \ref{fig:overview}), a dataset with graded, denser annotations for relationships between and among captions and images in the MS-COCO evaluation splits of \cite{karpathy2015deep} (with 25k English captions and 5k images each). 

CxC extends MS-COCO's existing image-caption pairs with continuous (0-5) semantic similarity ratings for those pairs and new pairs. The rating criteria extend those used for Semantic Textual Similarity \cite{agirre2012semeval}. Intramodal pairs are selected for annotation via an indirect sampling scheme biased to gain a broad distribution of similarities. In all, CxC contains \textit{human} ratings for \textbf{267,095} pairs (derived from 1,335,475 independent judgments), a massive extension in scale and detail to the 50k original binary pairings.

MS-COCO \textit{incompletely} supports three retrieval tasks: image-text, text-image and text-text. CxC enhances all of these with new positive pairs, and it also supports a \textit{new} image-image retrieval task. With its graded similarity judgements, CxC also supports correlation measures comparing model and human rankings. Retrieval metrics focus on positive pairs, but CxC's correlation scores additionally account for low-scoring items (non matches). Supporting these evaluations on a common set of images and captions makes them more valuable for understanding inter-modal learning---compared to disjoint sets of caption-image, caption-caption, and image-image associations. Also, multimodal representations such as CLIP \cite{radford2learning} are useful for downstream tasks such as Visual Question Answering \cite{Goyal_2017_CVPR}, Vision and Language Navigation \cite{majumdar2020vlnbert}, Referring Expressions \cite{yu2018mattnet} and Visual Commonsense Reasoning \cite{zellers2019recognition}, and we hope the additional relationships and evaluations provided by CxC will help develop even better representations for tasks that span these modalities.

To establish baselines for CxC, we provide results for existing unimodal models for text (Bag-of-Words, USE \cite{cer-etal-2018-universal}) and images (InceptionV3 \cite{szegedy2016rethinking}, ResNet-152 \cite{he2016deep}, SimCLRv2 \cite{DBLP:journals/corr/abs-2006-10029} as well as for two cross-modal retrieval models VSE++ \cite{DBLP:conf/bmvc/FaghriFKF18} and VSRN \cite{Li_2019_ICCV}. 


We furthermore demonstrate CxC's utility 
by evaluating a dual encoder that combines a bidirectional loss for image-text retrieval with a loss for text-text retrieval. The text encoder is composed of transformer layers over pre-trained BERT word representations and the image encoder is a pre-trained EfficientNet (B4) \cite{pmlr-v97-tan19a}.
This model delivers the strongest overall performance across all four retrieval tasks and correlation with human scores for text-text, image-image and image-text similarity. Compared to the same dual encoder trained only with image-text pairs, this model realizes small gains for image-text tasks and large gains for text-text task but with some degradation for image-image tasks. This indicates that the model trades capacity to encode images for better text encoding---an insight that would not be easily assessed without CxC's image-image annotations.



Our main contributions are the following:

\begin{itemize}[nosep]
    \item We describe a method for sampling items to get a broad distribution of similarities.
    \item We annotate the semantic similarity of 267,095 pairs. These enhance existing retrieval tasks and support a new image-image retrieval task. They also support correlation measures; these assess models' judgments of both positive \textit{and} negative associations.
    \item We establish baseline scores for existing models and a multitask dual encoder on all tasks and demonstrate that CxC allows model performance to be assessed more holistically.
    \item With its new positive pairs, CxC improves the recall@k measures common in image-text and text-image retrieval. This shows a 1-3\% increase in recall@k over several models.
    \item We release CxC's annotations at \url{https://github.com/google-research-datasets/Crisscrossed-Captions}, along with code to merge CxC with existing MS-COCO data.
\end{itemize}


\section{Dataset Collection}

\label{sec:collection}

Existing resources already support learning joint representations of images and text. However, we need better evaluation resources, so we extend the MS-COCO evaluation splits with graded similarity associations within and across modalities. MS-COCO has five captions for each image, split by \cite{karpathy2015deep} into 410k training, 25k development, and 25k test captions (82k/5k/5k for images). An ideal extension would rate every pair, but this is infeasible\footnote{A split with 5k images and 25k captions has $\approx$12.5M image-image, $\approx$312M caption-caption and $\approx$125M image-caption pairs, so annotating all items in the validation and test splits with 5 replications would require $\approx$4.5B judgments.} and most pairs are dissimilar anyway. To obtain new pairs with high expected similarity, we introduce a biased sampling scheme.

\begin{figure}
\begin{tabular}{l p{3.3cm}}
\subfloat{\includegraphics[align=t,width = 1.4in,keepaspectratio]{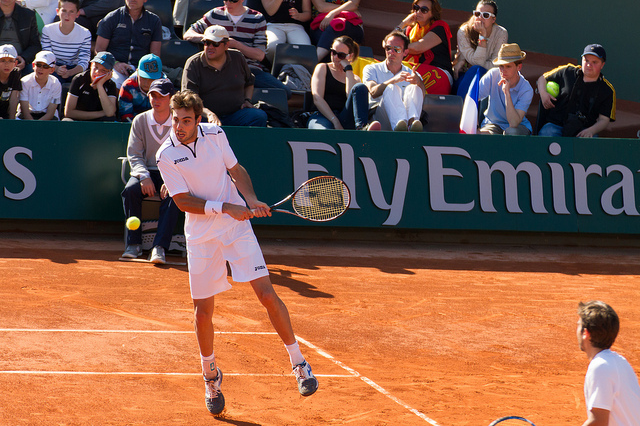}} &
\small Caption 1: A tennis player swinging a racket at a ball. \newline \newline
\small Caption 2: A man playing tennis with a crowd watching.  \\

\subfloat{\includegraphics[align=t,width = 1.4in,keepaspectratio]{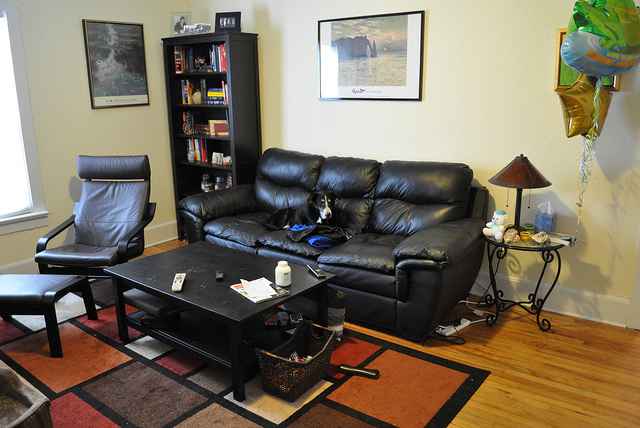}} &
\small Caption 3: A living room with some  black furniture and a colorful rug.  \newline \newline
\small Caption 4: A dog laying on a leather sofa in a living room.  \\ 
\end{tabular}
\caption{\textbf{Top:} Captions of the same image are often not paraphrases. \textbf{Bottom:} Such \textit{co-captions} often focus on different aspects and can diverge substantially.}
\label{image_caption_description}
\end{figure}

The data is collected in two phases. First, we define an indirect sampling scheme that uses model-based similarities from the co-modality items to select \textit{intra}modality pairs. We use these items and their human ratings to select \textit{inter}modality pairs for annotation. We also annotate all existing intermodal pairs and a large sample of co-captions (captions associated with the same image). See the appendix for details about the annotation interface and instructions, composition of the dataset and illustrative examples.

\begin{table*}
\small
\resizebox{16cm}{!}{
\begin{tabular}{p{6.6cm}p{7.2cm}ccc}
\textbf{Caption 1} & \textbf{Caption 2} & \textbf{STS} & \textbf{BoW} & \textbf{USE}  \\
\hline
A man standing on a  tennis court holding a racquet.   & A man standing on a tennis court holding a tennis racquet. & 5.0    & 0.98 & 0.99 \\
A man riding a skateboard off the side of a ramp. & A man riding up the side of a skateboard ramp.           & 4.2       & 0.99      & 0.94 \\
A yellow tray topped with a cup of coffee and a donut. & A white plate topped with donuts sitting on a stove top.   & 3.1 & 0.94 & 0.39 \\
A bird sitting on top of a park bench.            & An empty park bench sitting in front of trees.           & 2.2      & 0.90       & 0.69 \\
An old car sitting on top of a lush green field.  & A couple of motorcycles parked next to each other.       & 1.3      & 0.85      & 0.21 \\
A man sanding next to an orange frisbee.          & A couple of swans swimming in a pond next to two people. & 0.2      & 0.84      & 0.11 \\
\end{tabular}
}
\caption{A comparison of CxC STS annotation scores and cosine similarity scores using GloVe BoW embeddings and Universal Sentence Encoder (USE) for five example MS-COCO caption pairs.}
\label{tab:sts_bow_use_examples}
\end{table*}

\begin{figure*}
\centering
    \includegraphics[width=0.3\textwidth]{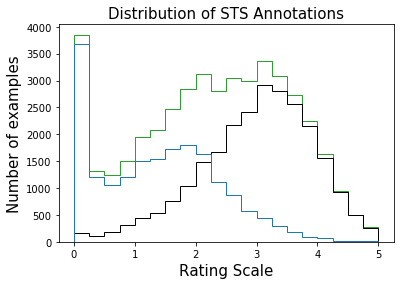}
    \hspace{15px}
    \includegraphics[width=0.3\textwidth]{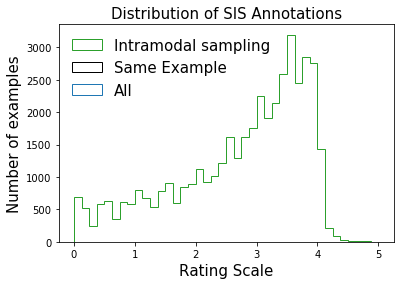}
    \hspace{15px}
    \includegraphics[width=0.3\textwidth]{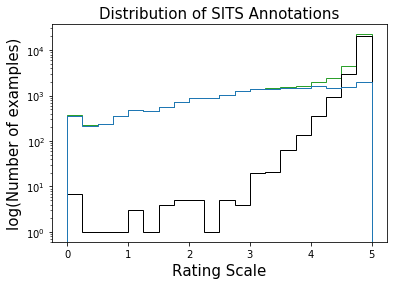}
    \caption{Distribution of ratings for the CxC validation set. \textit{Intramodal Sampling} refers to examples selected using \textit{other-modality} selection, \textit{Same Example} refers to original MS-COCO pairs, and \textit{All} covers all examples for a task.}
    \label{fig:ratings_val_plot}
\end{figure*}

\paragraph{Intramodality}
Two images of a man and a dog can be described differently, while two similar sentences about a man and a dog can describe dissimilar images. In Figure \ref{image_caption_description}, caption 1 gives a visual description while caption 2 gives a broader event description. Divergences also occur when caption creators perceive a scene differently: caption 3 describes the room and caption 4 focuses on the dog and sofa.
This semantic gap between images and their captions creates an opportunity to sample \textit{intra}modal pairs with varying similarities. Our key idea is to use model-based similarities of images for biased sampling of caption pairs, and vice versa, and use existing image-caption pairs as pivots between modalities. This selects image pairs that are different in appearance but similar in what they depict based on their descriptions, and vice versa. 

Denote the known images and captions as V ($v_1...v_n$) and C ($c_1...c_n$) (the latter representing \textit{co-caption} groups of five captions each). Each item is encoded with an off-the-shelf unimodal model. Cosine similarity between items defines two symmetric matrices: $S^C$ (pairwise caption similarities) and $S^V$ (pairwise image similarities). The diagonals are set to zero to not sample identical items.

We encode images with Graph-RISE \cite{48678} and construct $S^I$, the image-based similarity for pairs of co-caption groups. 
We encode captions with Universal Sentence Encoder (USE) \cite{cer-etal-2018-universal} and average bag of words (BoW) based on GloVe embeddings \cite{pennington-etal-2014-glove}. Co-caption representations are averaged to create a single representation. From these, we construct $S^C$, the caption-based similarity for images pairs. USE and BoW embeddings produce two $S^C$ matrices, but we gloss over this detail below.

We use $S^C$ to select \textit{image} pairs and $S^I$ for \textit{caption} pairs. Because of the cross-modal semantic gap, diversity and size of the underlying data, these pairs exhibit a wide range of similarity. Selecting the five most similar items (according to model-based $S^V$ and $S^C$) thus produces good representation of varying amounts of similarity as judged by people.
Because $S^V$ covers co-caption groups, one caption is randomly chosen from each group to produce a caption pair for rating.

Caption-caption and image-image candidates are referred to as $C2C$ and $I2I$, respectively. $I2I$ pairs are selected with the above \textit{other-modality} method. For $C2C$ pairs, we sample half the pairs using the other-modality method and half from within co-captions. The latter introduces (mostly) positive associations between caption pairs describing the same image. This gives a balanced set of caption pairs describing same and different images.



Pairs in $C2C$ and $I2I$ are scored by in-house raters using a continuous scale between 0 and 5. We adopt the widely used Semantic Textual Similarity (STS) \cite{cer-etal-2017-semeval} for text pairs and extend it to images to define Semantic Image Similarity (SIS). To recognize that this is a graded (rather than discrete) judgment, we encouraged raters to select scores like \textit{1.3} and obtain the final score for a pair as the average of five individual ratings.  


\paragraph{Intermodality}
We select caption-image candidates $C2I$ based on human ratings for $I2I$ and $C2C$ pairs. We mainly seek new positive matches like those identified by annotators in \citet{Ilharco2019LargescaleRL}. For each $I2I$ pair $(i_j,i_k)$, a $C2I$ pair $(c_k,i_j)$ is generated, where $c_k$ is a MS-COCO caption for $i_k$. We generate pairs from $C2C$ similarly. Half of the $C2I$ pairs are selected based on $C2C$ ranks and the other half by $I2I$ ranks (skipping pairs already selected from $C2C$). Finally, all MS-COCO pairs (25k in validation and 25k in test) are selected to obtain caption-image similarity ratings for the known items.


We extend STS to define Semantic Image-Text Similarity (SITS). Raters provide a continuous score from 0 to 5 using an interface similar to that for STS and SIS. Each $C2I$ pair receives five ratings; the average is used as the final SITS score. 


\begin{figure}
    \centering
    \includegraphics[width=0.4\textwidth]{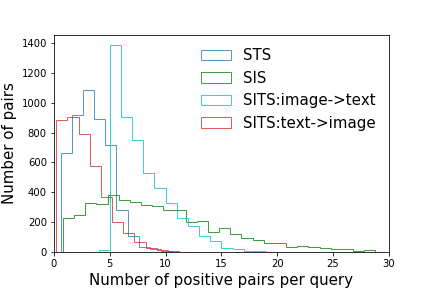}
    \caption{Distribution of counts of positive pairs (score $\geq$ 3) of annotations for each task (validation split).}
    \label{fig:query_val_plot}
\end{figure}

\section{Crisscrossed Captions Dataset}

Using our selection and annotation methodology, we obtained ratings for 267,095 caption-caption, image-image, and caption-image pairs (1,335,475 total judgments).
Figure \ref{fig:ratings_val_plot} shows rating distributions for each task (validation split). It also shows the distributions of ratings for STS and SIS pairs included from other-modality selection and from original MS-COCO pairs. The test set distributions are similar. Figure \ref{fig:query_val_plot} gives the distribution of counts of positive examples in each task (validation split), where a score $\geq 3$ (for STS, SITS) and a score $\geq 2.5$ (for SIS) is considered positive. These positive examples are used for intermodal and intramodal retrieval evaluation.

\begin{figure}
    \centering
    \includegraphics[width=0.4\textwidth]{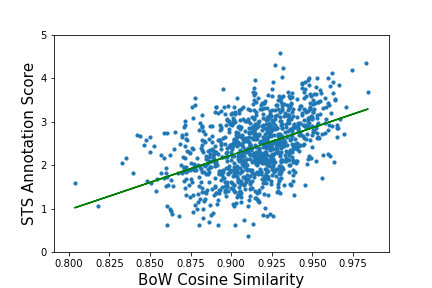}
    \caption{Plot BoW cosine similarity and STS scores for a sample of caption pairs, with the line of best fit.}
    \label{fig:sts_corr_plot}
\end{figure}

\textbf{STS.} The majority of caption pairs selected using image similarity are negative (ratings in [0, 3)), which is expected given the divergences noted in Figure \ref{image_caption_description}.  Nevertheless, the approach produces 20,587 positive pairs. Table \ref{tab:sts_bow_use_examples} shows pairs with their STS annotation scores and cosine similarity with BoW and USE embeddings. There is broad agreement, but the annotated similarity is not fully captured by either BoW or USE. USE provides a broader range, but scores the third pair lower than the fourth. BoW scores are bunched within a high similarity band\footnote{BoW scores fall mostly in the range .8 to 1.0 over all possible pairs; STS scores fall mostly in 1 (.2) to 4 (.8).} that aligns well with these five examples. Overall, there is a weak positive correlation between BoW and STS scores, as shown in Figure \ref{fig:sts_corr_plot}, which plots average BoW cosine similarity versus STS for 1000 randomly sampled pairs.

\begin{figure}
\includegraphics[width=\columnwidth]{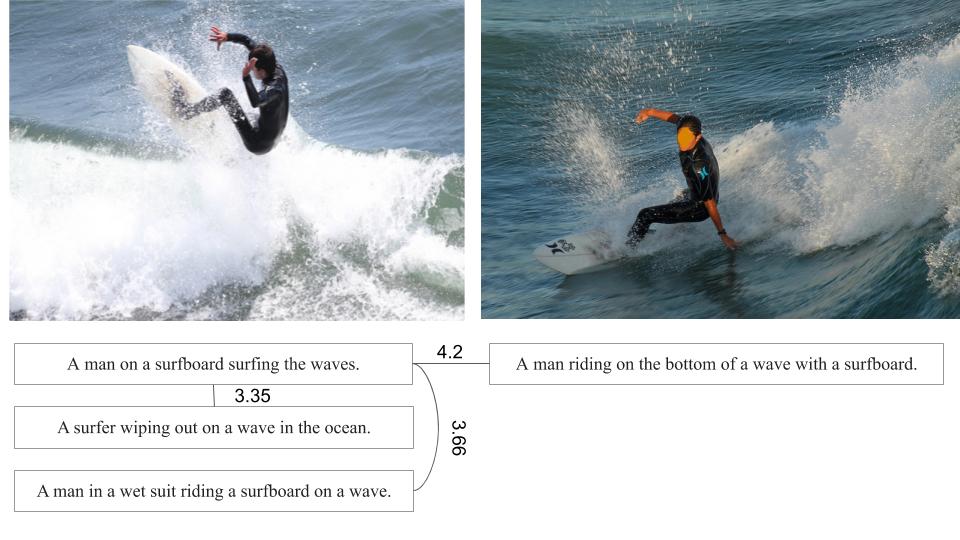}
\caption{An example of \textit{other-modality} caption pairs (horizontal pair) with higher STS compared to their respective co-captions (vertical pairs). Each caption is placed under its corresponding image.}
\label{fig:cocaption_vs_isim_pair}
\end{figure}

Figure \ref{fig:cocaption_vs_isim_pair} shows a pair of captions (and corresponding images) selected by the other-modality strategy with higher STS compared to their respective co-captions. For co-caption pairs, STS scores are more positive but many are still negative (Figure \ref{fig:ratings_val_plot}, left). Thus, combining both approaches leads to a more representative distribution overall. The large number of negative pairs from co-captions underscores the problem with assuming captions of the same image are paraphrases.

\textbf{SIS.} All image pairs $I2I$ are selected using the other-modality strategy. 
This plus the stringent criteria for SIS rating of 5 means very few examples are rated above 4.  Nevertheless, there are many pairs with SIS $\geq3$, indicating there are many images depicting similar scenes and events. 

\textbf{SITS.} As shown in Figure \ref{fig:query_val_plot}, there are many more pairs with 4-5 SITS ratings, compared to STS and SIS. This is by design, as the $C2I$ pairs are selected based on decreasing STS/SIS scores. This captures more positive intermodality associations and augments the existing validation and test splits. Since these pairs are missing from the existing data, they are among the examples that inappropriately penalize a model that identifies them correctly in image-caption retrieval. The SITS ratings collected for known pairs also support new correlation based evaluations, as discussed in the next section.

\section{Evaluation Tasks and Metrics}
CxC supports intermodal retrieval like MS-COCO but with denser annotations between image-caption pairs. It also enables intramodal retrieval and semantic similarity correlation evaluations, which were not possible before. 


\citet{karpathy2015deep} first used MS-COCO for image-to-caption and caption-to-image retrieval. We extend the existing associations with positive CxC pairs, and also add new caption-to-caption and image-to-image retrieval tasks using positive STS and SIS pairs (a total of four retrieval tasks). To the best of our best knowledge, CxC is the first dataset to support image-to-image retrieval over captioned images. Following \citet{karpathy2015deep}, we evaluate using Recall@K~(R@K), computed as the fraction of times a correct item was found among the top K results, and median rank (med. r) of the closest ground truth result in the list.

Semantic similarity tasks such as Semantic Textual Similarity~\cite{cer-etal-2017-semeval} and Visual Semantic Textual Similarity(vSTS)~\cite{vsts} require a model to produce a continuous similarity score given two inputs. Typically, the models are evaluated based on the Pearson's $r$ of their scores with the human judgments over a set of input pairs. This is valid when training data is available to calibrate model scores to the human ratings. With CxC, we do not have such training data, so we instead use Spearman's $r$ to assess whether a model \textit{ranks} pairs similarly to human raters. 

It would be tempting to simply measure Spearman's $r$ over all pairs, but this would be flawed because CxC's dense annotation means that the scores between many pairs are themselves correlated. To mitigate this, we use a sampled bootstrap correlation instead. For each correlation estimate, we sample half of the queries (to increase diversity across samples) and for each selected query, we choose one of the items for which CxC supplies a paired rating. We compute Spearman's $r$ between the CxC scores and the model scores for the selected pairs. The final correlation is the average over 1000 of these bootstrap samples. 

vSTS \cite{vsts} contains 2677 pairs of MS-COCO captions and corresponding images. As noted above, vSTS is related dataset for multimodal semantic similarity. We considered mixing CxC and vSTS; however, this was infeasible because CxC uses the widely adopted Karpathy splits, while items in vSTS’s training, dev and test splits are spread among the Karpathy splits. We could not just make a separate cut of CxC because vSTS pairs can cross splits, e.g. an image-caption item in Karpathy training and another in Karpathy test. Given the small size of vSTS, we focused our efforts on CxC evaluations.

\section{Evaluated Models}
In order to establish baselines for CxC, we benchmark pretrained models for images and text. Note that these are off-the-shelf models that have not been trained on MS-COCO. We also evaluate cross-modal retrieval models that are trained on MS-COCO. Here, we focus on models that support efficient retrieval (e.g. dual encoders). We expect models with extensive cross-modal interactions, such as ViLBERT \cite{NIPS2019_8297} and LXMERT \cite{tan-bansal-2019-lxmert}, will show strong performance on CxC tasks, either as standalone models that (inefficiently) score all possible item pairs or as rerankers for outputs of retrieval models.

To the best of our knowledge, there is no prior work that explores joint learning or evaluation on intra- and inter-modality retrieval tasks. \citet{ngiam2011multimodal} and \citet{collell2016image} show evidence that inter-modality learning helps improve intra-modality performance, but do not explore multitask learning. \citet{lu202012} explore multitask learning but only focus on intermodal representation learning for intermodal downstream tasks. To illustrate how CxC allows us to measure how intermodal representation learning can improve both intra- and inter-modal performance, we train a dual encoder model on bidirectional image-text and text-text in-batch retrieval losses.

\subsection{Pretrained Model Baselines}
\paragraph{Text-only Models} First, we use a bag-of-words (\textit{BoW}) approach using averaged GloVe embeddings~\cite{pennington-etal-2014-glove} for each token in a caption as the caption representation. Second, the Universal Sentence Encoder (\textit{USE})~\cite{cer-etal-2018-universal} is a sentence level representation model that has shown strong performance on the related STS benchmark. We use the multilingual transformer version from TensorFlow Hub~\cite{yang-etal-2020-multilingual}.\footnote{universal-sentence-encoder-multilingual-large/1}

\paragraph{Image-only Models} \textit{InceptionV3}, \textit{ResNet-152}, and \textit{SimCLRv2} are deep convolutional models ~\cite{szegedy2016rethinking, he2016deep, DBLP:journals/corr/abs-2002-05709, DBLP:journals/corr/abs-2006-10029} trained on the ImageNet dataset. We extract 2048-dimensional image-level representations on a central crop containing 87.5\% of the original image area.
We access them via TensorFlow Hub.\footnote{imagenet/inception\_v3/feature\_vector/4, imagenet/resnet\_v1\_152/feature\_vector/4 and gs://simclr-checkpoints/simclrv2/finetuned\_100pct/r50\_1x\_sk0/hub/ respectively}

\paragraph{Intermodal Models} \textit{VSE++}~\cite{DBLP:conf/bmvc/FaghriFKF18} is a dual encoder (see Sec. \ref{dual-encoder-models}) trained to learn a joint space of aligned images and captions. The state-of-the-art VSRN model \cite{Li_2019_ICCV} is another dual encoder that uses additional training annotations to predict and use bounding boxes for more fine-grained and coherent image analysis, while using only a simple text encoder trained from scratch.\footnote{Models available at https://github.com\/fartashf\/vsepp (checkpoint "runs/coco\_vse++/model\_best.pth.tar") and https://github.com/KunpengLi1994/VSRN (checkpoint "pretrain\_model/coco/model\_coco\_1.pth.tar").}

\begin{table*}
\centering
\small
\resizebox{0.8\textwidth}{!}{
\begin{tabular}{l@{\hspace{4mm}}l|cccc|cccc}
& & \multicolumn{4}{c|}{\textbf{Image $\rightarrow$ Text}} & \multicolumn{4}{c}{\textbf{Text $\rightarrow$ Image}} \\ 
{\bf Annotations} & {\bf Model} & R@1 & R@5 & R@10 & med r  & R@1 & R@5 & R@10 &  med r  \\ \hline
\multirow{4}{*}{\textbf{MS-COCO}} & VSE++ & 41.3 & 71.1 & 81.2 & 2  & 30.3 & 59.4 & 72.4 & 4  \\
& VSRN-github & 50.3 & 79.6 & 87.9 & 1 & 37.9 & 68.5 & 79.4 & 2 \\
& VSRN-paper & 53.0 & 81.1 & 89.4 & - & 40.5 & 70.6 & 81.1 & - \\
& \DEit & 52.0 & 80.3 & 89.5 & 1 & 37.9 & 67.6 & 78.8 & 2 \\
& \DEttit & 54.1 & 81.8 & 89.9 & 1 & 39.7 & 70.0 & 80.9 & 2 \\
[2mm] 
\multirow{4}{*}{\textbf{CxC}} & VSE++ & 43.1 & 74.3 & 84.2 & 2 & 32.5 & 62.7 & 75.4 & 3 \\
& VSRN-github & 52.4 & 81.9 & 90.0 & 1 & 40.1 & 71.1 & 81.5 & 2 \\
& \DEit & 53.9 & 82.7 & 91.2 & 1 & 39.8 & 70.2 & 80.9 & 2 \\
& \DEttit & 55.9 & 84.2 & 91.8 & 1 & 41.7 & 72.3 & 83.0 & 2 \\
\end{tabular}}
\caption{Image $\leftrightarrow$ Text retrieval results on the MS-COCO 5k test set and \textbf{CxC's extended pairs for the same}.}
\label{tab:inter_coco_vs_cxc_results}
\end{table*}

\subsection{Dual Encoder Baselines}
\label{dual-encoder-models}



We also consider several neural baseline models, all of which are \textit{dual encoders} \cite{gillick2018end,ijcai2019-746} that encode both 
inputs separately. 
Dual encoder models have been proven as an effective approach to learn strong semantic representations~\cite{cer-etal-2018-universal,DBLP:journals/corr/abs-2002-05709,DBLP:journals/corr/abs-2006-10029}. They are often trained using an \textit{in-batch sampled softmax} loss, as this has been observed to converge quickly and perform well on retrieval tasks \cite{gillick2018end,ijcai2019-746}.
We employ the bidirectional in-batch sampled softmax loss (eq. \ref{eq:rankloss}):
\begingroup
\setlength{\abovedisplayskip}{0pt}
\setlength{\belowdisplayskip}{12pt}
\begin{equation}
\small
\begin{aligned}
\mathcal{L} = -\frac{1}{K}  \sum_{i=1}^{K} \left( S(l_i, r_i) - \mathrm{log} \!\!\! \sum_{j=1\,j\neq{}i}^{K} \!\!\! e^{S(l_i,\,r_j)} \right) \\
-\frac{1}{K}  \sum_{i=1}^{K} \left( S(r_i, l_i) - \mathrm{log} \!\!\! \sum_{j=1\,j\neq{}i}^{K} \!\!\! e^{S(r_i,\,l_j)} \right)
\end{aligned}
\label{eq:rankloss}
\end{equation}
\endgroup

\noindent
where $S(x, y)$ is the dot product of embeddings of examples $x$ and $y$. This loss encourages the score of a correct pair  $S(l_i,\,r_i)$ to be higher than scores of non-matching input pairs from the batch $S(l_i,\,r_j)$.
Unlike full cross-attention models, this architecture enables large-scale retrieval through approximate nearest neighbor search. 

We train dual encoders for caption-image and caption-caption tasks, as well as a multitask model that combines both tasks. We use EfficientNet-B4 \cite{tan2019efficientnet} (pre-trained on ImageNet) as our image encoder; it yields a 1792-dimensional representation. The text encoder employs a frozen\footnote{Freezing BERT makes performance slightly worse, but makes training much faster.} BERT-Base model \cite{devlin-etal-2019-bert} followed by three transformer layers. The additional transformer layers have 8 attention heads, hidden dimension of 3072, and--like BERT-base--output 768-dimensional token-level features. We use the features at the $0^{th}$ token position of the final layer as the caption representation. BERT parameters are initialized from the public BERT checkpoint.\footnote{bert\_en\_uncased\_L-12\_H-768\_A-12/2} The additional, trainable transformer layers are randomly initialized.

We construct three dual encoder models from these base encoders. (1) A \textit{Text-Text} model (\DEtt) uses a shared text encoder for both sides. (2) An \textit{Image-Text} model (\DEit) uses the aforementioned text and image encoders, and includes a layer above the text encoder to project its 768 dimensions to 1792 (to match the image encoder output). (3) A \textit{Multitask} model (\DEttit) is trained on a combination of tasks ~\cite{chidambaram-etal-2019-learning}. It shares \DEit's architecture and is trained in the same way; however, its loss is a weighted sum of image-text ($i2t$, $t2i$) and text-text ($t2t$) losses:
\begin{equation}
    \mathcal{L} = \mathcal{L}_{i2t} + \mathcal{L}_{t2i} + c * \mathcal{L}_{t2t}
\end{equation}
Here $c$ is a scalar controlling the weights of losses from each task. This model has one text encoder, shared between all retrieval tasks. For hyperparameter tuning and training setup, see the appendix.

\section{Results}

\begin{table}
\centering
\small
\begin{tabular}{l|rrrr}
\multirow{2}{*}{\bf{Model}} & \multicolumn{4}{c}{\textbf{Text $\rightarrow$ Text}} \\ 
& R@1 & R@5 & R@10  & med r \\ \hline
BoW & 21.2 & 38.2 & 47.4 & 13 \\
USE\textsubscript{mling-large}   & 31.2 & 51.5 & 61.3 & 5 \\ 
VSE++ & 38.7 & 62.3 & 72.2 & 3 \\
VSRN-github & 41.0 & 64.8 & 74.5 & 2 \\
\DEtt & 41.7 & 64.4 & 73.4 & 2 \\
\DEit & 26.0 & 47.1 & 57.5 & 7 \\
\DEttit & 42.4 & 64.9 & 74.0 & 2 \\
\end{tabular}
\caption{Text $\leftrightarrow$ Text retrieval performance on MS-COCO 5k test set \textbf{using CxC annotations}.}
\label{tab:intra_Crisscrossed_results}
\end{table}

\paragraph{Intermodal Retrieval}
Table \ref{tab:inter_coco_vs_cxc_results} summarizes intermodal retrieval performance on both the original MS-COCO annotations and CxC. 
We report performance of two versions of VSRN \cite{Li_2019_ICCV}--one using the checkpoint on the author's Github (VSRN-github, which allows us to perform CxC evaluations) and the other from the original paper (VSRN-paper, which has higher MS-COCO scores).
Comparing each model on MS-COCO and CxC, the new positive items added by CxC show improved retrieval performance as they identify missing positives that are incorrectly penalized when using only original pairs (as noted in \citet{Ilharco2019LargescaleRL} for Flickr8k). Multitask training in \DEttit provides a boost over using only intermodal pairs for training (i.e. \DEit). It performs similarly with VSRN-paper---it seems likely that VSRN's greater investment on the image analysis (with representations based on extracted object bounding boxes) is matched by \DEttit's greater investment in the text encoder.

\begin{table}
\centering
\small
\begin{tabular}{l|rrrr}
\multirow{2}{*}{\bf{Model}} & \multicolumn{4}{c}{\textbf{Image $\rightarrow$ Image}} \\ 
      & R@1 & R@5 & R@10  & med r \\
\hline
InceptionV3             & 4.1 & 13.3 & 19.1 & 96  \\ 
ResNet-152              & 11.8 & 35.5 & 49.5 & 11 \\
SimCLRv2                & 24.5 & 54.9 & 68.1 & 4 \\ [1mm] 
VSE++                   & 36.4 & 70.4 & 81.3 & 2 \\
VSRN-github             & 44.2 & 76.7 & 86.2 & 2 \\
\DEit                   & 38.3 & 74.1 & 85.0 & 2 \\
\DEttit                 & 38.5 & 73.6 & 84.9 & 2 \\
\end{tabular}
\caption{Image $\leftrightarrow$ Image retrieval performance on MS-COCO 5k test set \textbf{using CxC annotations}.}
\label{tab:intra_Crisscrossed_results_i2i}
\end{table}

\begin{table}
\centering
\small
\begin{tabular}{l|rrr}
\multirow{2}{*}{\textbf{Model}} & \multicolumn{1}{c}{\textbf{STS}} & \multicolumn{1}{c}{\textbf{SIS}} & \multicolumn{1}{c}{\textbf{SITS}} \\
& avg $\pm$ std & avg $\pm$ std & avg $\pm$ std \\
\hline
BoW & 55.1$\pm$0.6 & -- & -- \\
USE\textsubscript{mling-large} & 71.4$\pm$0.4 & -- & --  \\
Inception V3 & -- & 19.6$\pm$1.9 & --  \\ 
ResNet-152 & --& 59.2$\pm$1.3 & --  \\ 
SimCLRv2 & -- & 74.3$\pm$0.9 & -- \\ [1mm] 
VSE++ & 74.4$\pm$0.4 & 73.3$\pm$0.9 & 55.2$\pm$1.5 \\
VSRN-github & 73.0$\pm$0.4 & 70.1$\pm$1.0 & 60.4$\pm$1.3 \\
\DEtt & 72.9$\pm$0.4 & -- & -- \\
\DEit & 50.9$\pm$0.6 & 81.3$\pm$0.7 & 61.6$\pm$1.4 \\
\DEttit & 74.2$\pm$0.4 & 74.5$\pm$0.9 & 61.9$\pm$1.3 \\
\end{tabular}
\caption{Spearman's R Bootstrap Correlation ($\times 100$) on MS-COCO 5k test set \textbf{using CxC annotations}.}
\label{spearman_corr}
\end{table}


\begin{figure}
\centering
\resizebox{\linewidth}{!}{%
\begin{tabular}{p{2.7cm}c}
\textbf{Caption} & \textbf{Ranked Images with CxC Score}  \\ 
    \multirow{2}{2.7cm}{\newline
    A person performs a stunt jump on a motorcycle.} &
    \subfloat{\includegraphics[align=t,width=1.1in, height=1.1in]{c2i_final_16_0.png} 
    \hspace{1px}
    \includegraphics[align=t,width=1.1in, height=1.1in]{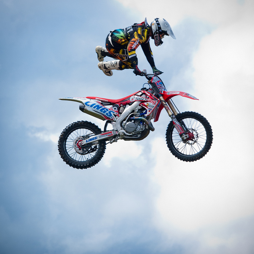}
    \hspace{1px}
    \includegraphics[align=t,width=1.1in, height=1.1in]{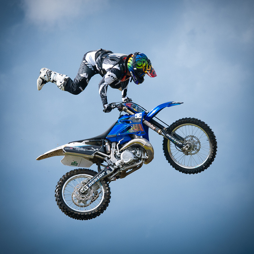}} \\
    & 5.0  \quad\quad\quad\quad\quad\quad 4.95 \textbf{*} \quad\quad\quad\quad\quad\quad 4.45 \\
    
    \multirow{2}{2.7cm}{\newline \newline
    A bear is holding on to a rock by some water.} &
    \subfloat{\includegraphics[align=t,width=1.1in, height=1.1in]{c2i_final_14_0.png}
    \hspace{1px}
    \includegraphics[align=t,width=1.1in, height=1.1in]{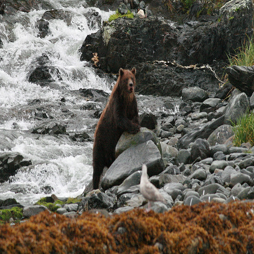}
    \hspace{1px}
    \includegraphics[align=t,width=1.1in, height=1.1in]{c2i_final_14_2.png}} \\
    & N/A \quad\quad\quad\quad\quad\quad 4.83 \textbf{*} \quad\quad\quad\quad\quad\quad 4.29  \\ 
    
    \multirow{2}{2.7cm}{\newline \newline 
    An old street looking sign made of wood.} &
    \subfloat{\includegraphics[align=t,width=1.1in, height=1.1in]{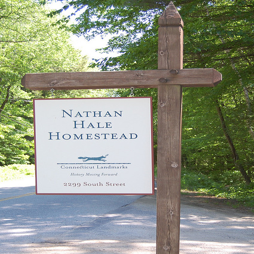}
    \hspace{1px}
    \includegraphics[align=t,width=1.1in, height=1.1in]{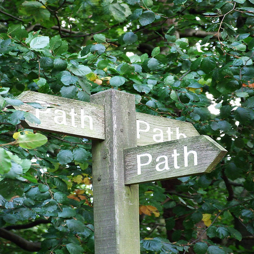}
    \hspace{1px}
    \includegraphics[align=t,width=1.1in, height=1.1in]{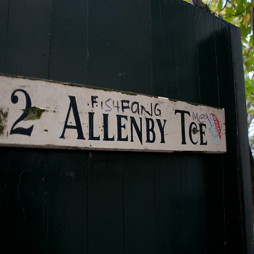}} \\
    & 5.0 \quad\quad\quad\quad\quad\quad 4.98 \textbf{*} \quad\quad\quad\quad\quad\quad N/A \\
    
\end{tabular}}
\caption{Text{$\rightarrow$}Image retrieval examples (MS-COCO val set), with CxC SITS score provided when available. Images are ranked from left to right, based on \DEttit scores. MS-COCO annotations only consider images marked by \textbf{*} to be correct retrieval results.
}
\label{fig:t2i examples}
\end{figure}

\begin{figure}
\centering
\resizebox{\linewidth}{!}{%
\begin{tabular}{lp{11cm}} 
\textbf{Image} & \textbf{(CxC score) Ranked Captions} \\ 
    \subfloat{\includegraphics[align=t,width=1in]{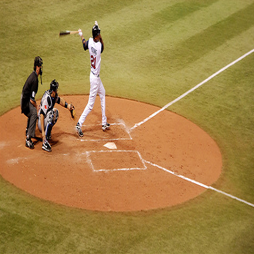}}&
    (4.48) Home plate at a professional baseball game, batter not quite ready. \newline
    (4.46) three players on the base ball diamond, all headed for a base. \newline
    (4.15) Baseball team mates and another player on the diamond. \newline
    (4.98) A batter, catcher and umpire in a baseball game. \newline
    (4.95) A batter, catcher and umpire in a  baseball game. \newline \\

  \subfloat{\includegraphics[align=t,width=1in, height=1in]{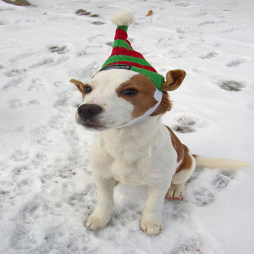}} &
    (4.92) \textbf{A dog wearing a striped elf hat sits in the snow.} \newline
    (5.0) \textbf{A dog is wearing an elf hat in the snow.} \newline
    (5.0) \textbf{A dog wearing an elf hat sits in the snow.} \newline
    (4.25) \textbf{Brown and white dog in Christmas hat standing in the snow.} \newline
    (4.98) A dog that is wearing a christmas hat on its head. \newline \\
    
    \subfloat{\includegraphics[align=t,width=1in, height=1in, keepaspectratio]{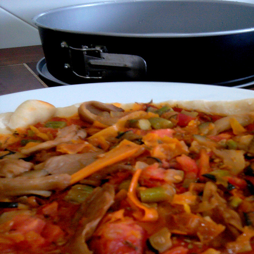}} &
    (4.75) \textbf{A plate of food with peppers, onions and meats.} \newline
    (5.0) A pot of vegetables is cooking on a stove. \newline
    (4.61) Chicken cordon blue and fries with a garnish. \newline
    (4.9) A pot with some food in it on a stove. \newline
    (4.52) \textbf{some kind of noodle and vegetable dish being made on the stove.} \newline \\
\end{tabular}}
\caption{Image{$\rightarrow$}Text retrieval results (MS-COCO val set), ranked top to bottom using \DEttit scores. MS-COCO annotations only consider the \textbf{bold} captions to be correct results.}
\label{fig:i2t examples}
\end{figure}

Figure \ref{fig:t2i examples} shows three examples of images retrieved for caption queries. The CxC annotations capture missing examples in the first two cases, and the last two show there are still more positive pairs that remain unassociated in CxC. Figure \ref{fig:i2t examples} shows the same for captions retrieved from image queries, again showing that many examples are captured in CxC that are missing in MS-COCO. 

\paragraph{Intramodal Retrieval}
Tables \ref{tab:intra_Crisscrossed_results} and \ref{tab:intra_Crisscrossed_results_i2i} give intramodal retrieval results enabled by CxC's STS and SIS ratings respectively. USE\textsubscript{mling-large}  is a strong baseline for Text{$\rightarrow$}Text, but all the cross-modal models beat USE\textsubscript{mling-large} by a wide margin, likely due to learning on in-domain captions. InceptionV3 and ResNet-152 prove surprisingly weak for Image{$\rightarrow$}Image, but SimCLRv2 proves to be a strong unimodal baseline for this task. The cross-modal models nevertheless beat SimCLRv2 by a wide margin, even though none were trained on image-image retrieval directly. In terms of joint intra- and inter-modal learning, the multitask \DEttit model provides strong, balanced performance: it is close to \DEtt for Text{$\rightarrow$}Text and \DEit for Image{$\rightarrow$}Image and far outperforms the latter for Text{$\rightarrow$}Text. The strong performance is especially notable considering that both \DEit and \DEttit have the same model capacity.

\paragraph{Semantic Similarity}
Table \ref{spearman_corr} shows Spearman's R bootstrapped correlation for all models with respect to CxC's STS, SIS and SITS scores. Overall, VSE++, VSRN-github and \DEttit perform better than unimodal baselines, but interesting further patterns emerge. Despite being much worse for retrieval, VSE++ actually beats VSRN-github on STS and SIS; however, its low SITS score indicates it fails to bridge the two modalities as well. The correlation scores also show that \DEit is too focused on images: it has the highest SIS (81.3), but has worse STS (50.9) than even BoW (55.1). Adding the text-text loss to \DEit training, i.e. \DEttit, produces much more balanced overall performance. On SIS, SimCLRv2 is stronger than all cross-modal models, except \DEit. SITS scores appear to rank all models similarly to retrieval (Table \ref{tab:inter_coco_vs_cxc_results}).


The fact that \DEttit is better than both \DEtt and unimodal baselines for STS and Text{$\rightarrow$}Text retrieval is encouraging, and it demonstrates the value of having a single set of annotations covering the relatedness of a common set of images and captions. We expect that a multitask model which also uses image-image training pairs could demonstrate gains across all tasks---measurements made possible by the CxC annotations (especially the new image-image associations).

\section{Conclusion}
The CxC dataset provides a much more complete set of relationships between and among images and captions than the raw MS-COCO image-caption pairs. We demonstrate that a dual encoder that learns from both image-caption pairs and caption-caption pairs (\DEttit) exhibits strong, balanced performance across four retrieval tasks and three correlation measures. There is much remaining headroom for future models for all these tasks.

CxC's annotations themselves validate the strong semantic alignment between images and their original captions---these have an average similarity of 4.85. However, we also find that co-captions (captions for the same image) have an average score of just 3.0. This calls into question the use of such pairs in training and evaluating paraphrase generation models \cite{Gupta2018ADG} and reinforces the need for images as context for human evaluation in paraphrasing \cite{DBLP:conf/aaai/WangGCB19}.  


\section*{Acknowledgments}
We thank Julia Hockenmaier for her inputs on CxC's formulation, the Google Data Compute Team, especially Ashwin Kakarla and Mohd Majeed for their tooling and annotation support, Yuan Zhang, Eugene Ie for their comments on the initial versions of the paper and Daphne Luong for executive support for the data collection.

\bibliography{anthology,eacl2021}
\bibliographystyle{acl_natbib}

\include{appendix}

\end{document}

%% file: appendix.tex
\subsection*{CxC Annotations - Collection and Analysis}
\setcounter{figure}{0}
\setcounter{table}{0}


We present additional details of the human annotation process.
To annotate the CxC dataset, in house annotators were employed: 170 (74 men, 96 women) for STS, 61 (28 men, 33 women) for SIS and 113 (46 men, 67 women) for SITS. All were aged between 20-35 years.
The annotators were paid hourly wages that are competitive for their locale. They have standard rights as contractors. They were fluent English speakers.

We define separate annotation interfaces for each of Semantic Textual Similarity (STS), Semantic Image Similarity (SIS) and Semantic Image Text Similarity (SITS) tasks.
We define a similarity scale ranging from 0 to 5 for all three tasks, following ~\citet{cer-etal-2017-semeval}.

\begin{figure*}[b!]
    \centering
    \includegraphics[width=0.96\textwidth]{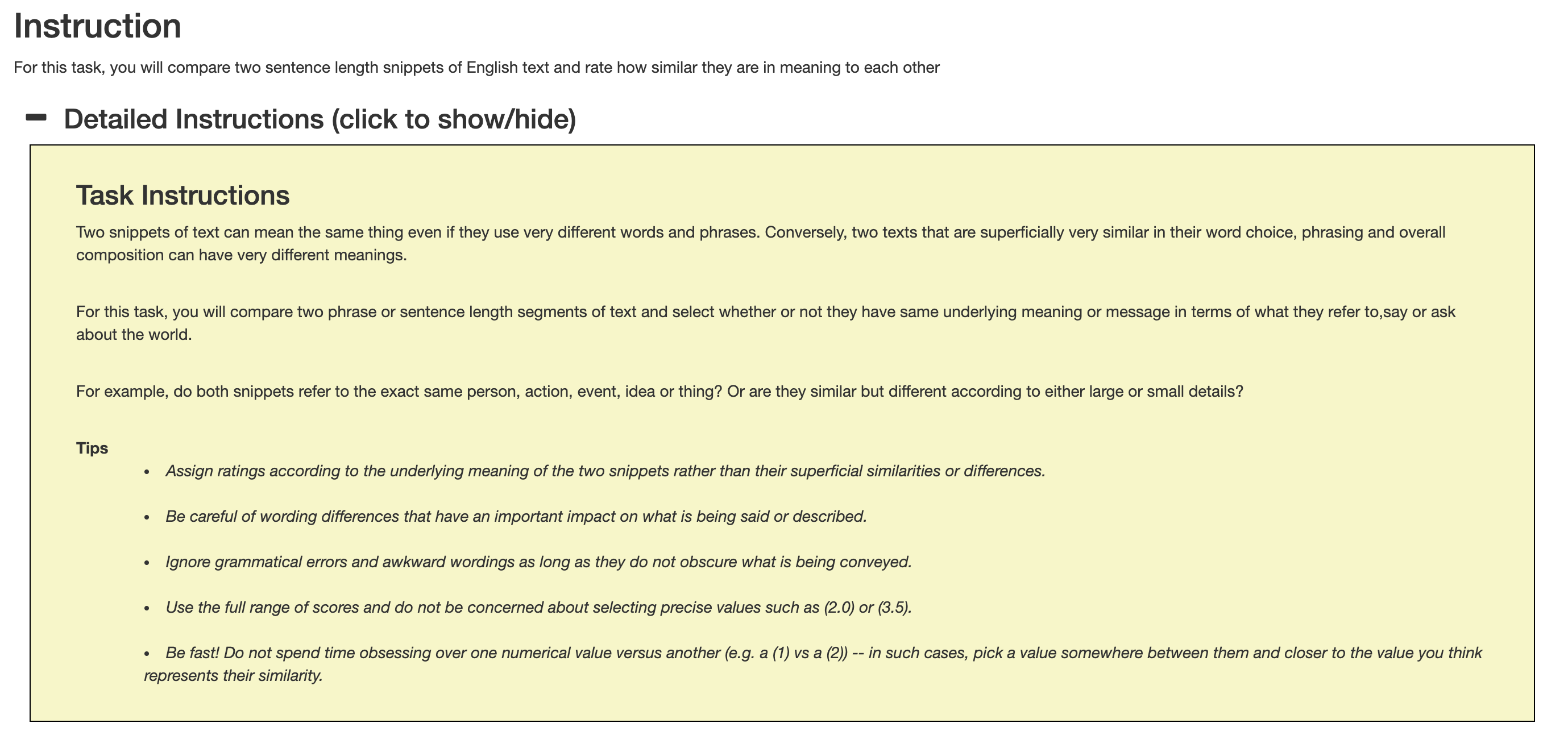}
    \includegraphics[width=0.96\textwidth]{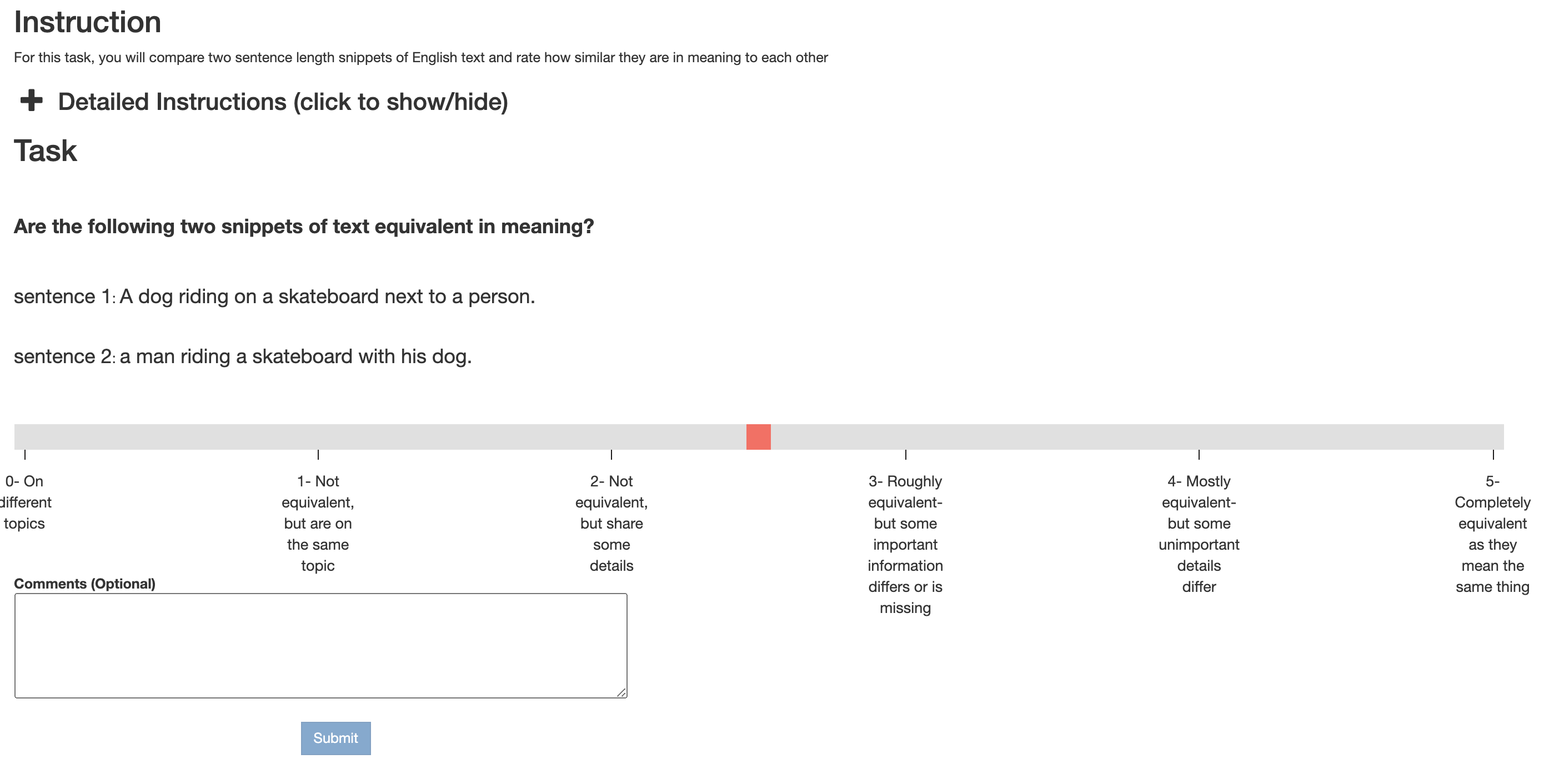}
    \caption{STS Annotation Interface}
    \label{fig:sts_ann_ui}
\end{figure*}

\begin{figure*}
    \centering
    \includegraphics[height=0.46\textheight]{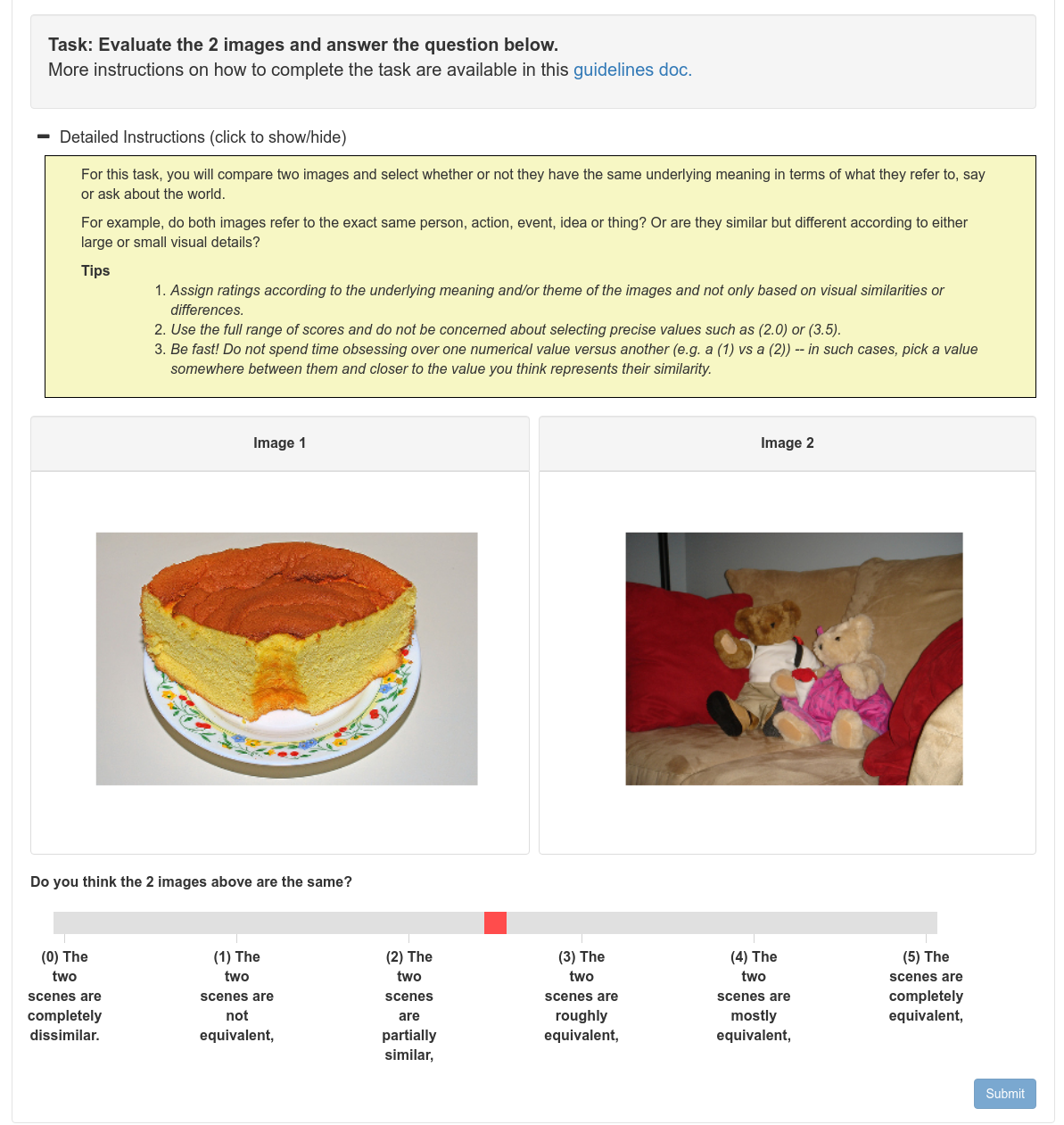}
    \caption{SIS Annotation Interface}
    \label{fig:sis_ann_ui}
\end{figure*}

\begin{figure*}
    \centering
    \includegraphics[height=0.46\textheight]{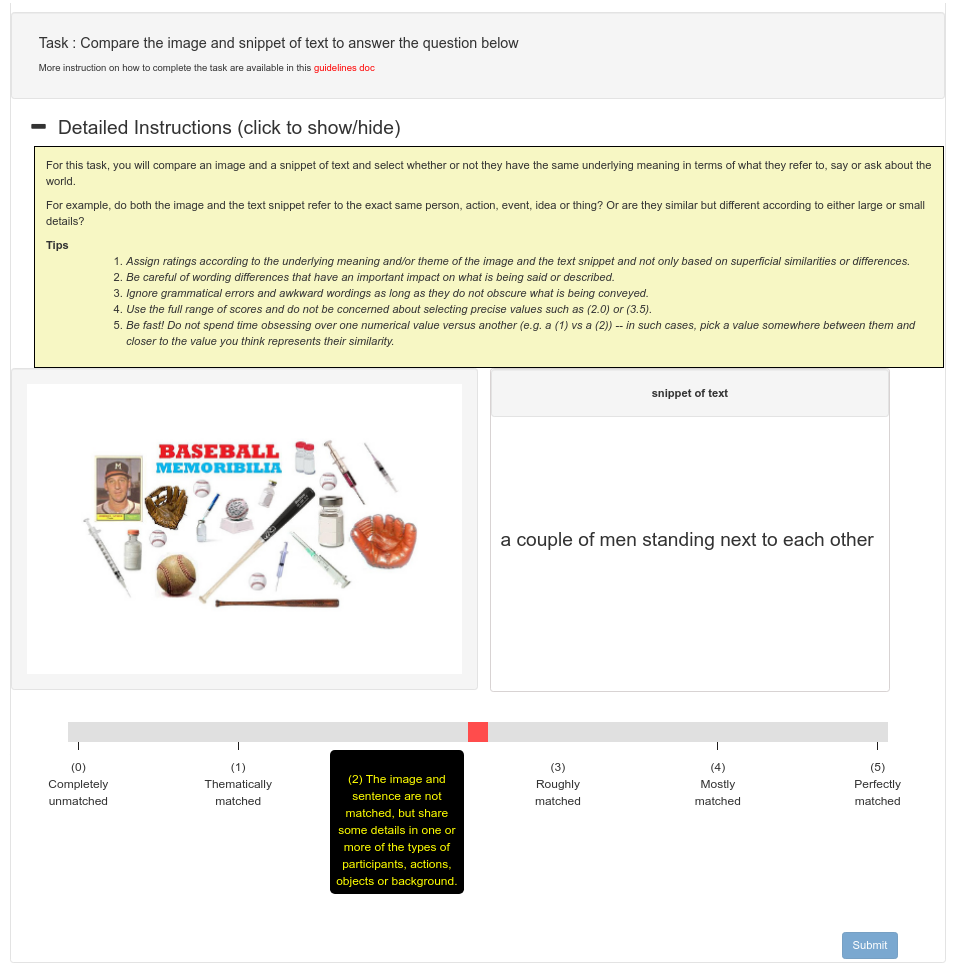}
    \caption{SITS Annotation Interface}
    \label{fig:sits_ann_ui}
\end{figure*}

We conducted a few pilot annotation rounds with the annotators to evaluate the effectiveness of the annotation instructions and the annotation interface. 
We learned that allowing the annotators to rate on a continuous 0-5 scale instead of a discrete one like STS resulted in higher correlation between the individual ratings. 
As a result, we decided to use the continuous ratings in the task but still keep the similarity definition for each discrete value in the annotation instructions.
The final annotation interfaces are illustrated in Figures \ref{fig:sts_ann_ui} for STS, \ref{fig:sis_ann_ui} for SIS and \ref{fig:sits_ann_ui} for SITS.
Task-specific high-level instructions are displayed at the top in an expandable text box followed by a pair of examples. At the bottom there is a sliding bar with 0-5 score instructions along the scale. Since the SITS instructions are longer, they are shown when the annotator hovers over the corresponding score to improve readability for this task.

Each annotator is required to evaluate the displayed example based on the instructions and score them. The sliding scale makes it intuitive for the annotators to rate an example 2.87 if they feel the semantic similarity of the pair lies between score descriptions of 2 and 3, leaning towards 3. Finally, the annotator response is recorded when they click the submit button at the bottom of the page. The absolute score is deliberately not displayed so as not to distract the workers towards trying to get a clean integer value like 3.0 instead of 2.94 or 2.97. 

\begin{figure}
\centering
\resizebox{\linewidth}{!}{%
\begin{tabular}{c | l l || l p{3cm}}
\small
  & \multicolumn{2}{c||}{ \textbf{Image Pairs~-~SIS annotations}}& \multicolumn{2}{c}{ \textbf{Image Text Pairs~-~SITS annotations.}}\\ \hline
 5 &
\subfloat{\includegraphics[align=t,width = 1.2in,height=1.2in]{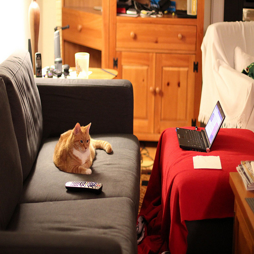}} &
\subfloat{\includegraphics[align=t,width = 1.2in,height=1.2in]{sis_rating_5_1.png}} &
\subfloat{\includegraphics[align=t,width = 1.2in,height=1.2in]{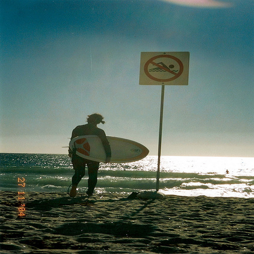}} &
 A man poses with a surfboard on a beach.  \\ 

 4 &
\subfloat{\includegraphics[align=t,width = 1.2in,height=1.2in,keepaspectratio]{sis_rating_4_0.png}} &
\subfloat{\includegraphics[align=t,width = 1.2in,height=1.2in,keepaspectratio]{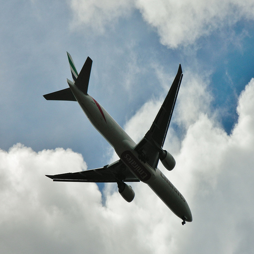}} &
\subfloat{\includegraphics[align=t,width = 1.2in,height=1.2in,keepaspectratio]{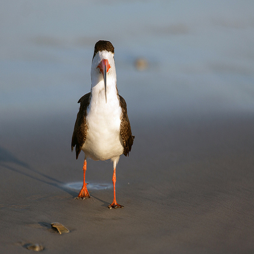}} &
 A couple of birds that are walking on some sand.  \\

 3 &
\subfloat{\includegraphics[align=t,width = 1.2in,height=1.2in,keepaspectratio]{sis_rating_3_0.png}} &
\subfloat{\includegraphics[align=t,width = 1.2in,height=1.2in,keepaspectratio]{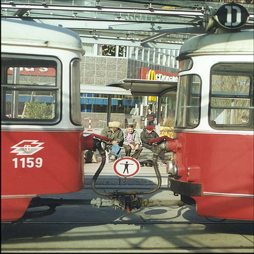}} &
\subfloat{\includegraphics[align=t,width = 1.2in,height=1.2in,keepaspectratio]{sits_rating_3.png}} &
 A man is riding a surfboard at the beach.  \\  

 2 &
\subfloat{\includegraphics[align=t,width = 1.2in,height=1.2in,keepaspectratio]{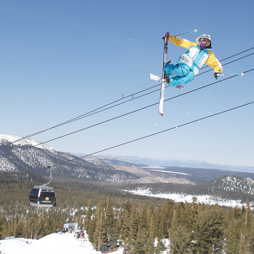}} &
\subfloat{\includegraphics[align=t,width = 1.2in,height=1.2in,keepaspectratio]{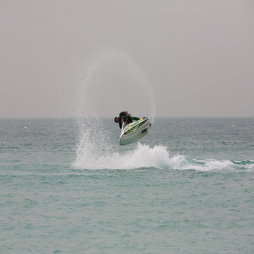}} &
\subfloat{\includegraphics[align=t,width = 1.2in,height=1.2in,keepaspectratio]{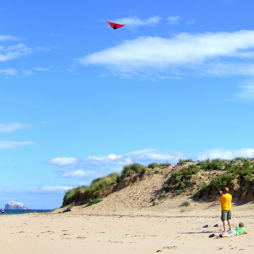}} &
 Three people stand on an empty beach watching a bird in the sky. \\  

 1 &
\subfloat{\includegraphics[align=t,width = 1.2in,height=1.2in,keepaspectratio]{sis_rating_1_0.png}} &
\subfloat{\includegraphics[align=t,width = 1.2in,height=1.2in,keepaspectratio]{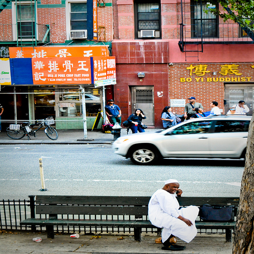}} &
\subfloat{\includegraphics[align=t,width = 1.2in,height=1.2in,keepaspectratio]{sits_rating_1.png}} &
 A man in a hat rides an elephant in a river.  \\  

 0 &
\subfloat{\includegraphics[align=t,width = 1.2in,height=1.2in,keepaspectratio]{sis_rating_0_0.png}} &
\subfloat{\includegraphics[align=t,width = 1.2in,height=1.2in,keepaspectratio]{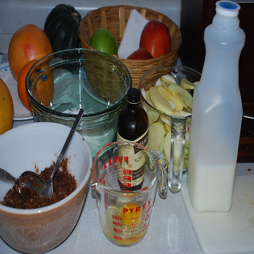}} &
\subfloat{\includegraphics[align=t,width = 1.2in,height=1.2in,keepaspectratio]{sits_rating_0.png}} &
 Long road with a sign titled  Jackson River Rd and East Main St. \\   

\end{tabular}}
\caption{Examples for each annotation score (0-5) of SIS (left) and SITS (right) tasks.}
\label{fig:sis_sits_examples}
\end{figure}

The annotators were able to get a better grasp of the task through the pilot annotations and got quicker at scoring the pairs. They took an average of 37, 17 and 17 seconds per example for STS, SIS and SITS tasks respectively for the final round of annotations. Table \ref{sts_sis_rating_criteria} describes the instructions shared with the annotation workers for each task. The side-by-side comparison shows how each rating on the SIS and SITS scales compares to the STS benchmark. Figure \ref{fig:sis_sits_examples} shows a set of SIS and SITS examples for the 0-5 rating scale shared along with the instructions. Table \ref{tab:ratings_dist_table} contains the breakdown of the number of annotations per task per split. 

\begin{table}
\resizebox{\columnwidth}{!}{%
\begin{tabular}{l|rrrc}
\backslashbox{Split}{Task}  & STS   & SIS   & SITS  & Total (per split) \\ \hline
Validation       & 44,009 & 42,767 & 44,722 & 131,498          \\ 
Test             & 44,045 & 46,719 & 44,833 & 135,597         \\ 
Total (per task) & 88,054 & 89,486 & 89,555 & 267,095         \\ 
\end{tabular}
}

\caption{Number of annotations per task and split.}
\label{tab:ratings_dist_table}
\end{table}

\begin{figure*}
    \centering
    \includegraphics[width=\linewidth]{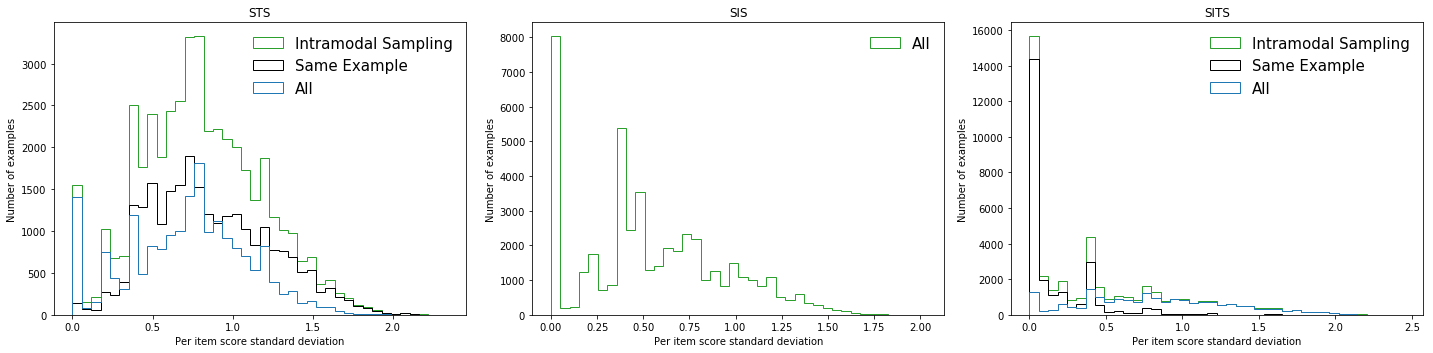}
    \caption{Standard deviation per item in CxC. \textit{Intramodal Sampling} refers to examples selected using \textit{other-modality} selection, \textit{Same Example} refers to original MS-COCO pairs, and \textit{All} covers all examples for a task.}
    \label{fig:per_item_vairance}
\end{figure*}

Figure \ref{fig:per_item_vairance} shows a distribution of the standard deviation of raw annotations for each item per task. For STS, there is larger overall deviation compared to the other two tasks--it seems that pairs of short captions leave more ambiguity and are open for broader interpretation than when at least one image is involved. Note also that SITS is expected to have lower deviation because of the sampling based on STS and SIS annotations. 


\subsection*{Training Setup}
\begin{figure}    
\centering
    \includegraphics[width=0.6\columnwidth]{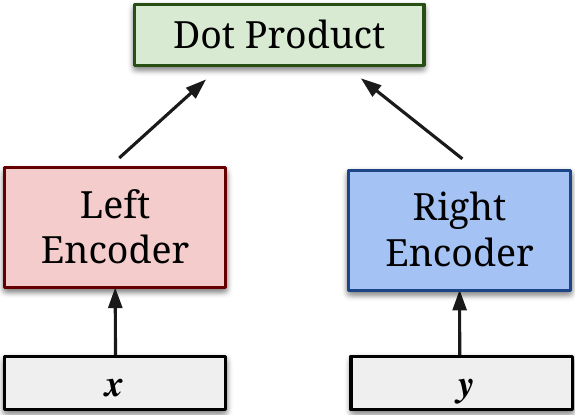}
    \caption{Dual Encoder with inputs $x$ and $y$, encoded by the left and right encoders, respectively. Similarity is computed as the dot product of the encodings.}
    \label{fig:dual_encoder}
\end{figure}

Figure \ref{fig:dual_encoder} shows the basic architecture of the dual encoder models from Section 5.2, which establish strong baselines on all the retrieval and correlation tasks.  The image encoder and text encoder are both pre-trained in all experiments.
Following \newcite{Ilharco2019LargescaleRL}, we pretrain our dual encoders on the Conceptual Captions dataset \cite{sharma2018conceptual} with image-to-caption and caption-to-image losses.
Conceptual Captions contains 3.3 million pairs of images and captions---far larger than MS-COCO. Pre-training uses the Adam optimizer ($\beta_1=0.9, \beta_2=0.999$) and a learning rate that starts at 1e-4 and decays by 0.1\% every 1000 steps. We stop pre-training after $\approx$30k steps and select the checkpoint that maximizes R@10 on a held-out set. We then fine-tune this checkpoint on MS-COCO using the same hyper parameters, except for a smaller learning rate of 5e-6.

Our models are trained on 32-core slices of Cloud TPU V3 pods, with a per-replica batch size of $K=64$ during both pre-training and fine-tuning. Because in-batch sampled softmax loss is known to perform best when computed over a large number of negative samples \cite{gillick2018end}, our training setup pools image and caption encodings from all replicas before computing the loss. That is, each replica computes $l$ and $r$ for its local mini-batch and broadcasts them to all others to be used as negative samples. Training with $N$ cores thus allows the loss to be computed over the global batch of $N{\cdot}K$ examples and $(N{\cdot}K)^2$ pairs (in our case 2048 examples and $2048^2$ example pairs).

\begin{table*}
{\small
\begin{tabular}{c|p{3.5cm}|p{5.2cm}|p{5.3cm}}
Score &  Semantic Textual Similarity (STS) & Semantic Image Similarity (SIS) & Semantic Image-Text Similarity (SITS)\\
\hline
5 & \textbf{The texts are completely equivalent} as they mean the same thing. & \textbf{The scenes are near duplicates}, possibly being  viewed from a different perspective. & \textbf{The image and sentence are perfectly matched.} The sentence is an almost perfect description for the image. \\ 
4 & \textbf{The texts are mostly equivalent} but some unimportant details differ. & \textbf{The two scenes are mostly equivalent}, but some unimportant details differ such as involving  different but the same or highly similar types of participants, actions, objects and background. & \textbf{The image and sentence are mostly matched}, but some unimportant details differ such as involving different but the same or highly similar types of participants, actions, objects and background. The text can partially describe the image. \\
3 & \textbf{The texts are roughly equivalent} but some important information differs or is missing. & \textbf{The two scenes are roughly equivalent}, but some important details are different or missing such as involving a notable difference in the types  of participants, actions, objects or background. & \textbf{The image and sentence are roughly matched}, but some important details are different or missing such as involving a notable difference in the types of participants, actions, objects or background. The image cannot be described using the text. \\
2 &  \textbf{The texts are not equivalent} but share some details. & \textbf{The two scenes are not equivalent}, but share some details in terms of the types of participants, actions, objects or background. & \textbf{The image and sentence are not matched}, but share some details in one or more of the types of participants, actions, objects or background. \\
1 & \textbf{The texts are not equivalent} but are on the same topic. & \textbf{The two scenes are not equivalent}, but are loosely thematically related. & \textbf{The image and sentence are not matched}, but are loosely thematically related. \\ 
0 & \textbf{The texts are on different topics.} & \textbf{The two scenes are completely dissimilar.} & \textbf{The image and sentence are completely unmatched.}\\ 
\end{tabular}
}
\caption{Intramodality annotation criteria for Semantic Image Similarity (SIS) and Intermodality annotation criteria for Semantic Image-Text Similarity (SITS) with comparison to equivalent Semantic Textual Similarity (STS) annotations \cite{agirre2012semeval}.}
\label{sts_sis_rating_criteria}
\end{table*}

\subsection*{Ablation Experiments}
Our model architecture and training setup differ from prior work in key ways. 
In particular, best known results for VSE++ and VSRN are from models that were trained with much smaller batch sizes, did not undergo Conceptual Captions pre-training, and had different image encoder architectures. To evaluate the effect of these factors, we trained variants of our \DEit model (here, the \emph{baseline} training recipe) with the following one-off ablations:
\begin{itemize}
  \item The \emph{small batch size} ablation reduces the training batch size to 128 examples, to match that of VSE++ and VSRN in \cite{DBLP:conf/bmvc/FaghriFKF18} and \cite{Li_2019_ICCV}, respectively.
  \item \emph{No pretraining} skips dual encoder pretraining on Conceptual Captions.
  \item \emph{ResNet-152} uses the same recipe as \emph{baseline}, but replaces the EfficientNet-B4 image encoder with ResNet-152, which was used in VSE++. Notably, EfficientNet-B4 has fewer parameters than ResNet-152, but achieves higher classification accuracy on ImageNet.
\end{itemize}

Table \ref{tab:de_hparam_ablation} summarizes the performance of the ablated models.
Reducing the batch size causes a small but consistent reduction in recalls across all tasks. Removing Conceptual Captions pretraining leads to larger regressions on all tasks -- except on Text-Text retrieval, where results are curiously better than the baseline. Likewise, models using ResNet-152 image encoders perform worst overall, but also perform (slightly) better than the baseline on Text-Text retrieval.

Overall, we conclude that pretraining and choice of image encoder architecture have large effects on model performance; large-batch training is beneficial, but has a smaller impact. Finally, the asymmetric shifts in task performance suggest models make implicit trade-offs based on the relative difficulty of each task -- here, apparently, a function of encoder strength and quantity of training data.  Understanding these dynamics, and building models that perform well across all tasks, requires future study. Crisscrossed Captions enables such work by giving a more complete picture of model quality on both intra- and inter-modal tasks.

\begin{table*}
\centering
\small
\resizebox{\textwidth}{!}{
\begin{tabular}{l@{\hspace{4mm}}l|cc|cc|cc|cc}
& & \multicolumn{2}{c|}{\textbf{Image $\rightarrow$ Text}} & \multicolumn{2}{c}{\textbf{Text $\rightarrow$ Image}} & \multicolumn{2}{c}{\textbf{Text $\rightarrow$ Text}} &
\multicolumn{2}{c}{\textbf{Image $\rightarrow$ Image}}\\ 
{\bf Annotations} & {\bf Model} & R@1 & R@10 & R@1 & R@10 & R@1 & R@10 & R@1 & R@10  \\ \hline
\multirow{4}{*}{\textbf{MS-COCO}} 
& \DEit (baseline) & 52.0 & 89.5 & 37.9 & 78.8 & 25.9 & 57.2 & -- & -- \\
& \DEit (small batch size) & 49.6 & 88.2 & 35.7 & 78.0 & 25.3 & 57.6 & -- & -- \\
& \DEit (no pretraining) & 45.0 & 86.0 & 31.2 & 74.7 & 34.5 & 67.2 & -- & -- \\
& \DEit (ResNet-152) & 43.5	& 83.0 & 28.9 & 71.4 & 28.2 & 60.2 & -- & --  \\
[2mm] 
\multirow{4}{*}{\textbf{CxC}}
& \DEit (baseline) & 53.9 & 91.2 & 39.8 & 80.9 & 26.0 & 57.5 & 38.3 & 85.0  \\
& \DEit (small batch size) & 51.8 & 90.1 & 37.7 & 80.3 & 25.4 & 57.9 & 38.0 & 84.3 \\
& \DEit (no pretraining) & 47.0 & 88.1 & 33.2 & 77.6 & 34.6 & 67.6 & 37.0 & 84.0  \\
& \DEit (ResNet-152) & 45.2	& 85.1 & 30.8 & 74.4 & 28.3 & 60.5 & 29.7 & 76.5 \\
\end{tabular}}
\caption{Ablation analysis for \DEit,  retrieval results on MS-COCO 5k test set and CxC. (Note that MS-COCO does not support Image $\rightarrow$ Image retrieval evaluation at all.)}
\label{tab:de_hparam_ablation}
\end{table*}


%% file: eacl2021.bbl
\begin{thebibliography}{41}
\expandafter\ifx\csname natexlab\endcsname\relax\def\natexlab#1{#1}\fi

\bibitem[{486()}]{48678}


\bibitem[{Agirre et~al.(2012)Agirre, Cer, Diab, and
  Gonzalez-Agirre}]{agirre2012semeval}
Eneko Agirre, Daniel Cer, Mona Diab, and Aitor Gonzalez-Agirre. 2012.
\newblock Semeval-2012 task 6: A pilot on semantic textual similarity.
\newblock In \emph{* SEM 2012: The First Joint Conference on Lexical and
  Computational Semantics--Volume 1: Proceedings of the main conference and the
  shared task, and Volume 2: Proceedings of the Sixth International Workshop on
  Semantic Evaluation (SemEval 2012)}, pages 385--393.

\bibitem[{Cer et~al.(2017)Cer, Diab, Agirre, Lopez-Gazpio, and
  Specia}]{cer-etal-2017-semeval}
Daniel Cer, Mona Diab, Eneko Agirre, I{\~n}igo Lopez-Gazpio, and Lucia Specia.
  2017.
\newblock \href {https://doi.org/10.18653/v1/S17-2001} {{S}em{E}val-2017 task
  1: Semantic textual similarity multilingual and crosslingual focused
  evaluation}.
\newblock In \emph{Proceedings of the 11th International Workshop on Semantic
  Evaluation ({S}em{E}val-2017)}, pages 1--14, Vancouver, Canada. Association
  for Computational Linguistics.

\bibitem[{Cer et~al.(2018)Cer, Yang, Kong, Hua, Limtiaco, St.~John, Constant,
  Guajardo-Cespedes, Yuan, Tar, Strope, and Kurzweil}]{cer-etal-2018-universal}
Daniel Cer, Yinfei Yang, Sheng-yi Kong, Nan Hua, Nicole Limtiaco, Rhomni
  St.~John, Noah Constant, Mario Guajardo-Cespedes, Steve Yuan, Chris Tar,
  Brian Strope, and Ray Kurzweil. 2018.
\newblock \href {https://doi.org/10.18653/v1/D18-2029} {Universal sentence
  encoder for {E}nglish}.
\newblock In \emph{Proceedings of the 2018 Conference on Empirical Methods in
  Natural Language Processing: System Demonstrations}, pages 169--174,
  Brussels, Belgium. Association for Computational Linguistics.

\bibitem[{Chen et~al.(2020{\natexlab{a}})Chen, Kornblith, Norouzi, and
  Hinton}]{DBLP:journals/corr/abs-2002-05709}
Ting Chen, Simon Kornblith, Mohammad Norouzi, and Geoffrey~E. Hinton.
  2020{\natexlab{a}}.
\newblock \href {https://arxiv.org/abs/2002.05709} {A simple framework for
  contrastive learning of visual representations}.
\newblock \emph{CoRR}, abs/2002.05709.

\bibitem[{Chen et~al.(2020{\natexlab{b}})Chen, Kornblith, Swersky, Norouzi, and
  Hinton}]{DBLP:journals/corr/abs-2006-10029}
Ting Chen, Simon Kornblith, Kevin Swersky, Mohammad Norouzi, and Geoffrey~E.
  Hinton. 2020{\natexlab{b}}.
\newblock \href {https://arxiv.org/abs/2006.10029} {Big self-supervised models
  are strong semi-supervised learners}.
\newblock \emph{CoRR}, abs/2006.10029.

\bibitem[{Chidambaram et~al.(2019)Chidambaram, Yang, Cer, Yuan, Sung, Strope,
  and Kurzweil}]{chidambaram-etal-2019-learning}
Muthu Chidambaram, Yinfei Yang, Daniel Cer, Steve Yuan, Yunhsuan Sung, Brian
  Strope, and Ray Kurzweil. 2019.
\newblock \href {https://doi.org/10.18653/v1/W19-4330} {Learning cross-lingual
  sentence representations via a multi-task dual-encoder model}.
\newblock In \emph{Proceedings of the 4th Workshop on Representation Learning
  for NLP (RepL4NLP-2019)}, pages 250--259, Florence, Italy. Association for
  Computational Linguistics.

\bibitem[{Collell~Talleda and Moens(2016)}]{collell2016image}
Guillem Collell~Talleda and Marie-Francine Moens. 2016.
\newblock Is an image worth more than a thousand words? on the fine-grain
  semantic differences between visual and linguistic representations.
\newblock In \emph{Proceedings of the 26th International Conference on
  Computational Linguistics}, pages 2807--2817. ACL.

\bibitem[{Devlin et~al.(2019)Devlin, Chang, Lee, and
  Toutanova}]{devlin-etal-2019-bert}
Jacob Devlin, Ming-Wei Chang, Kenton Lee, and Kristina Toutanova. 2019.
\newblock \href {https://doi.org/10.18653/v1/N19-1423} {{BERT}: Pre-training of
  deep bidirectional transformers for language understanding}.
\newblock In \emph{Proceedings of the 2019 Conference of the North {A}merican
  Chapter of the Association for Computational Linguistics: Human Language
  Technologies, Volume 1 (Long and Short Papers)}, pages 4171--4186,
  Minneapolis, Minnesota. Association for Computational Linguistics.

\bibitem[{Elliott et~al.(2016)Elliott, Frank, Sima{'}an, and
  Specia}]{elliott-etal-2016-multi30k}
Desmond Elliott, Stella Frank, Khalil Sima{'}an, and Lucia Specia. 2016.
\newblock \href {https://doi.org/10.18653/v1/W16-3210} {{M}ulti30{K}:
  Multilingual {E}nglish-{G}erman image descriptions}.
\newblock In \emph{Proceedings of the 5th Workshop on Vision and Language},
  pages 70--74, Berlin, Germany. Association for Computational Linguistics.

\bibitem[{Faghri et~al.(2018)Faghri, Fleet, Kiros, and
  Fidler}]{DBLP:conf/bmvc/FaghriFKF18}
Fartash Faghri, David~J. Fleet, Jamie~Ryan Kiros, and Sanja Fidler. 2018.
\newblock \href {http://bmvc2018.org/contents/papers/0344.pdf} {{VSE++:}
  improving visual-semantic embeddings with hard negatives}.
\newblock In \emph{British Machine Vision Conference 2018, {BMVC} 2018,
  Newcastle, UK, September 3-6, 2018}, page~12. {BMVA} Press.

\bibitem[{Gillick et~al.(2018)Gillick, Presta, and Tomar}]{gillick2018end}
Daniel Gillick, Alessandro Presta, and Gaurav~Singh Tomar. 2018.
\newblock End-to-end retrieval in continuous space.
\newblock \emph{arXiv preprint arXiv:1811.08008}.

\bibitem[{Goyal et~al.(2017)Goyal, Khot, Summers-Stay, Batra, and
  Parikh}]{Goyal_2017_CVPR}
Yash Goyal, Tejas Khot, Douglas Summers-Stay, Dhruv Batra, and Devi Parikh.
  2017.
\newblock Making the v in vqa matter: Elevating the role of image understanding
  in visual question answering.
\newblock In \emph{Proceedings of the IEEE Conference on Computer Vision and
  Pattern Recognition (CVPR)}.

\bibitem[{Gupta et~al.(2018)Gupta, Agarwal, Singh, and Rai}]{Gupta2018ADG}
Ankush Gupta, Arvind Agarwal, Prawaan Singh, and Piyush Rai. 2018.
\newblock A deep generative framework for paraphrase generation.
\newblock In \emph{Proceedings of AAAI-18}.

\bibitem[{Harwath and Glass(2017)}]{harwath-glass-2017-learning}
David Harwath and James Glass. 2017.
\newblock \href {https://doi.org/10.18653/v1/P17-1047} {Learning word-like
  units from joint audio-visual analysis}.
\newblock In \emph{Proceedings of the 55th Annual Meeting of the Association
  for Computational Linguistics (Volume 1: Long Papers)}, pages 506--517,
  Vancouver, Canada. Association for Computational Linguistics.

\bibitem[{He et~al.(2016)He, Zhang, Ren, and Sun}]{he2016deep}
Kaiming He, Xiangyu Zhang, Shaoqing Ren, and Jian Sun. 2016.
\newblock Deep residual learning for image recognition.
\newblock In \emph{Proceedings of the IEEE conference on computer vision and
  pattern recognition}, pages 770--778.

\bibitem[{Ilharco et~al.(2019)Ilharco, Zhang, and
  Baldridge}]{Ilharco2019LargescaleRL}
Gabriel Ilharco, Yuan Zhang, and Jason Baldridge. 2019.
\newblock \href {https://doi.org/10.18653/v1/K19-1006} {Large-scale
  representation learning from visually grounded untranscribed speech}.
\newblock In \emph{Proceedings of the 23rd Conference on Computational Natural
  Language Learning (CoNLL)}, pages 55--65, Hong Kong, China. Association for
  Computational Linguistics.

\bibitem[{Juan et~al.(2020)Juan, Lu, Li, Peng, Timofeev, Chen, Gao, Duerig,
  Tomkins, and Ravi}]{graph-rise}
Da-Cheng Juan, Chun-Ta Lu, Zhen Li, Futang Peng, Aleksei Timofeev, Yi-Ting
  Chen, Yaxi Gao, Tom Duerig, Andrew Tomkins, and Sujith Ravi. 2020.
\newblock \href {https://doi.org/10.1145/3336191.3371784} {Ultra fine-grained
  image semantic embedding}.
\newblock In \emph{Proceedings of the 13th International Conference on Web
  Search and Data Mining}, WSDM '20, page 277–285, New York, NY, USA.
  Association for Computing Machinery.

\bibitem[{Karpathy and Li(2015)}]{karpathy2015deep}
Andrej Karpathy and Fei-Fei Li. 2015.
\newblock Deep visual-semantic alignments for generating image descriptions.
\newblock In \emph{Proceedings of the IEEE conference on computer vision and
  pattern recognition}, pages 3128--3137.

\bibitem[{Kiros et~al.(2018)Kiros, Chan, and Hinton}]{kiros2018illustrative}
Jamie Kiros, William Chan, and Geoffrey Hinton. 2018.
\newblock Illustrative language understanding: Large-scale visual grounding
  with image search.
\newblock In \emph{Proceedings of the 56th Annual Meeting of the Association
  for Computational Linguistics (Volume 1: Long Papers)}, pages 922--933.

\bibitem[{de~Lacalle et~al.(2020)de~Lacalle, Salaberria, Soroa, Azkune, and
  Agirre}]{vsts}
Oier~Lopez de~Lacalle, Ander Salaberria, Aitor Soroa, Gorka Azkune, and Eneko
  Agirre. 2020.
\newblock Evaluating multimodal representations on visual semantic textual
  similarity.
\newblock In \emph{Proceedings of the Twenty-third European Conference on
  Artificial Intelligence (ECAI)}, Santiago Compostela, Spain.

\bibitem[{Li et~al.(2019)Li, Zhang, Li, Li, and Fu}]{Li_2019_ICCV}
Kunpeng Li, Yulun Zhang, Kai Li, Yuanyuan Li, and Yun Fu. 2019.
\newblock Visual semantic reasoning for image-text matching.
\newblock In \emph{The IEEE International Conference on Computer Vision
  (ICCV)}.

\bibitem[{Lin et~al.(2014)Lin, Maire, Belongie, Hays, Perona, Ramanan,
  Doll{\'a}r, and Zitnick}]{lin2014microsoft}
Tsung-Yi Lin, Michael Maire, Serge Belongie, James Hays, Pietro Perona, Deva
  Ramanan, Piotr Doll{\'a}r, and C~Lawrence Zitnick. 2014.
\newblock Microsoft coco: Common objects in context.
\newblock In \emph{European conference on computer vision}, pages 740--755.
  Springer.

\bibitem[{Lu et~al.(2019)Lu, Batra, Parikh, and Lee}]{NIPS2019_8297}
Jiasen Lu, Dhruv Batra, Devi Parikh, and Stefan Lee. 2019.
\newblock \href
  {http://papers.nips.cc/paper/8297-vilbert-pretraining-task-agnostic-visiolinguistic-representations-for-vision-and-language-tasks.pdf}
  {Vilbert: Pretraining task-agnostic visiolinguistic representations for
  vision-and-language tasks}.
\newblock In H.~Wallach, H.~Larochelle, A.~Beygelzimer, F.~d\textquotesingle
  Alch\'{e}-Buc, E.~Fox, and R.~Garnett, editors, \emph{Advances in Neural
  Information Processing Systems 32}, pages 13--23. Curran Associates, Inc.

\bibitem[{Lu et~al.(2020)Lu, Goswami, Rohrbach, Parikh, and Lee}]{lu202012}
Jiasen Lu, Vedanuj Goswami, Marcus Rohrbach, Devi Parikh, and Stefan Lee. 2020.
\newblock 12-in-1: Multi-task vision and language representation learning.
\newblock In \emph{Proceedings of the IEEE/CVF Conference on Computer Vision
  and Pattern Recognition}, pages 10437--10446.

\bibitem[{Majumdar et~al.(2020)Majumdar, Shrivastava, Lee, Anderson, Parikh,
  and Batra}]{majumdar2020vlnbert}
Arjun Majumdar, Ayush Shrivastava, Stefan Lee, Peter Anderson, Devi Parikh, and
  Dhruv Batra. 2020.
\newblock \href {http://arxiv.org/abs/2004.14973} {Improving
  vision-and-language navigation with image-text pairs from the web}.

\bibitem[{Ngiam et~al.(2011)Ngiam, Khosla, Kim, Nam, Lee, and
  Ng}]{ngiam2011multimodal}
Jiquan Ngiam, Aditya Khosla, Mingyu Kim, Juhan Nam, Honglak Lee, and Andrew~Y
  Ng. 2011.
\newblock Multimodal deep learning.
\newblock In \emph{ICML}.

\bibitem[{Pennington et~al.(2014)Pennington, Socher, and
  Manning}]{pennington-etal-2014-glove}
Jeffrey Pennington, Richard Socher, and Christopher Manning. 2014.
\newblock \href {https://doi.org/10.3115/v1/D14-1162} {{G}love: Global vectors
  for word representation}.
\newblock In \emph{Proceedings of the 2014 Conference on Empirical Methods in
  Natural Language Processing ({EMNLP})}, pages 1532--1543, Doha, Qatar.
  Association for Computational Linguistics.

\bibitem[{Radford et~al.()Radford, Kim, Hallacy, Ramesh, Goh, Agarwal, Sastry,
  Askell, Mishkin, Clark et~al.}]{radford2learning}
Alec Radford, Jong~Wook Kim, Chris Hallacy, Aditya Ramesh, Gabriel Goh,
  Sandhini Agarwal, Girish Sastry, Amanda Askell, Pamela Mishkin, Jack Clark,
  et~al.
\newblock Learning transferable visual models from natural language
  supervision.
\newblock \emph{Image}, 2:T2.

\bibitem[{Rashtchian et~al.(2010)Rashtchian, Young, Hodosh, and
  Hockenmaier}]{rashtchian2010collecting}
Cyrus Rashtchian, Peter Young, Micah Hodosh, and Julia Hockenmaier. 2010.
\newblock Collecting image annotations using amazon's mechanical turk.
\newblock In \emph{Proceedings of the NAACL HLT 2010 Workshop on Creating
  Speech and Language Data with Amazon's Mechanical Turk}, pages 139--147.
  Association for Computational Linguistics.

\bibitem[{Sharma et~al.(2018)Sharma, Ding, Goodman, and
  Soricut}]{sharma2018conceptual}
Piyush Sharma, Nan Ding, Sebastian Goodman, and Radu Soricut. 2018.
\newblock Conceptual captions: A cleaned, hypernymed, image alt-text dataset
  for automatic image captioning.
\newblock In \emph{Proceedings of ACL}.

\bibitem[{Szegedy et~al.(2016)Szegedy, Vanhoucke, Ioffe, Shlens, and
  Wojna}]{szegedy2016rethinking}
Christian Szegedy, Vincent Vanhoucke, Sergey Ioffe, Jon Shlens, and Zbigniew
  Wojna. 2016.
\newblock Rethinking the inception architecture for computer vision.
\newblock In \emph{Proceedings of the IEEE conference on computer vision and
  pattern recognition}, pages 2818--2826.

\bibitem[{Tan and Bansal(2019)}]{tan-bansal-2019-lxmert}
Hao Tan and Mohit Bansal. 2019.
\newblock \href {https://doi.org/10.18653/v1/D19-1514} {{LXMERT}: Learning
  cross-modality encoder representations from transformers}.
\newblock In \emph{Proceedings of the 2019 Conference on Empirical Methods in
  Natural Language Processing and the 9th International Joint Conference on
  Natural Language Processing (EMNLP-IJCNLP)}, pages 5100--5111, Hong Kong,
  China. Association for Computational Linguistics.

\bibitem[{Tan and Le(2019{\natexlab{a}})}]{pmlr-v97-tan19a}
Mingxing Tan and Quoc Le. 2019{\natexlab{a}}.
\newblock \href {http://proceedings.mlr.press/v97/tan19a.html}
  {{E}fficient{N}et: Rethinking model scaling for convolutional neural
  networks}.
\newblock volume~97 of \emph{Proceedings of Machine Learning Research}, pages
  6105--6114, Long Beach, California, USA. PMLR.

\bibitem[{Tan and Le(2019{\natexlab{b}})}]{tan2019efficientnet}
Mingxing Tan and Quoc Le. 2019{\natexlab{b}}.
\newblock Efficientnet: Rethinking model scaling for convolutional neural
  networks.
\newblock In \emph{International Conference on Machine Learning}, pages
  6105--6114.

\bibitem[{Wang et~al.(2019)Wang, Gupta, Chang, and
  Baldridge}]{DBLP:conf/aaai/WangGCB19}
Su~Wang, Rahul Gupta, Nancy Chang, and Jason Baldridge. 2019.
\newblock \href {https://doi.org/10.1609/aaai.v33i01.33017176} {A task in a
  suit and a tie: Paraphrase generation with semantic augmentation}.
\newblock In \emph{The Thirty-Third {AAAI} Conference on Artificial
  Intelligence, {AAAI} 2019, The Thirty-First Innovative Applications of
  Artificial Intelligence Conference, {IAAI} 2019, The Ninth {AAAI} Symposium
  on Educational Advances in Artificial Intelligence, {EAAI} 2019, Honolulu,
  Hawaii, USA, January 27 - February 1, 2019}, pages 7176--7183. {AAAI} Press.

\bibitem[{Yang et~al.(2020)Yang, Cer, Ahmad, Guo, Law, Constant, Abrego, Yuan,
  Tar, Sung, Strope, and Kurzweil}]{yang-etal-2020-multilingual}
Yinfei Yang, Daniel Cer, Amin Ahmad, Mandy Guo, Jax Law, Noah Constant,
  Gustavo~Hernandez Abrego, Steve Yuan, Chris Tar, Yun-hsuan Sung, Brian
  Strope, and Ray Kurzweil. 2020.
\newblock \href {https://doi.org/10.18653/v1/2020.acl-demos.12} {Multilingual
  universal sentence encoder for semantic retrieval}.
\newblock In \emph{Proceedings of the 58th Annual Meeting of the Association
  for Computational Linguistics: System Demonstrations}, pages 87--94, Online.
  Association for Computational Linguistics.

\bibitem[{Yang et~al.(2019)Yang, Hernandez~Abrego, Yuan, Guo, Shen, Cer, Sung,
  Strope, and Kurzweil}]{ijcai2019-746}
Yinfei Yang, Gustavo Hernandez~Abrego, Steve Yuan, Mandy Guo, Qinlan Shen,
  Daniel Cer, Yun-hsuan Sung, Brian Strope, and Ray Kurzweil. 2019.
\newblock \href {https://doi.org/10.24963/ijcai.2019/746} {Improving
  multilingual sentence embedding using bi-directional dual encoder with
  additive margin softmax}.
\newblock In \emph{Proceedings of the Twenty-Eighth International Joint
  Conference on Artificial Intelligence, {IJCAI-19}}, pages 5370--5378.
  International Joint Conferences on Artificial Intelligence Organization.

\bibitem[{Young et~al.(2014)Young, Lai, Hodosh, and
  Hockenmaier}]{young2014image}
Peter Young, Alice Lai, Micah Hodosh, and Julia Hockenmaier. 2014.
\newblock From image descriptions to visual denotations: New similarity metrics
  for semantic inference over event descriptions.
\newblock \emph{Transactions of the Association for Computational Linguistics},
  2:67--78.

\bibitem[{Yu et~al.(2018)Yu, Lin, Shen, Yang, Lu, Bansal, and
  Berg}]{yu2018mattnet}
Licheng Yu, Zhe Lin, Xiaohui Shen, Jimei Yang, Xin Lu, Mohit Bansal, and
  Tamara~L Berg. 2018.
\newblock Mattnet: Modular attention network for referring expression
  comprehension.
\newblock In \emph{Proceedings of the IEEE Conference on Computer Vision and
  Pattern Recognition}, pages 1307--1315.

\bibitem[{Zellers et~al.(2019)Zellers, Bisk, Farhadi, and
  Choi}]{zellers2019recognition}
Rowan Zellers, Yonatan Bisk, Ali Farhadi, and Yejin Choi. 2019.
\newblock From recognition to cognition: Visual commonsense reasoning.
\newblock In \emph{Proceedings of the IEEE Conference on Computer Vision and
  Pattern Recognition}, pages 6720--6731.

\end{thebibliography}
